\documentclass[a4paper,fleqn]{cas-dc}
\usepackage[justification=centering]{caption}
\usepackage{graphicx} 
\usepackage{subfigure}
\usepackage{lineno,hyperref}
\usepackage{float}
\usepackage{cas-common}
\usepackage[numbers]{natbib}

\usepackage{algorithm}
\usepackage{algorithmicx}
\usepackage{color}
\usepackage{algpseudocode}
\usepackage{amsmath}
\begin{document}
\captionsetup[figure]{labelfont={bf},name={Fig.},labelsep=period}
\let\WriteBookmarks\relax
\def\floatpagepagefraction{1}
\def\textpagefraction{.001}
\shorttitle{A graph convolution-based generative shilling attack}
\shortauthors{Fan Wu et~al.}

\title [mode = title]{Ready for Emerging Threats to Recommender Systems? A Graph Convolution-based Generative Shilling Attack}

\author[1,2]{Fan Wu}
\address[1]{Key Laboratory of Dependable Service Computing in Cyber Physical Society (Chongqing University), Ministry of Education, Chongqing, 401331, China}
\address[2]{School of Big Data and Software Engineering, Chongqing University, Chongqing, 401331, China}

\author[1,2]{Min Gao}
\cormark[1]
\ead{gaomin@cqu.edu.cn}

\author[3]{ Junliang Yu}
\address[3]{School of Information Technology and Electrical Engineering, The University of Queensland, Queensland, 4072, Australia}

\author[2]{Zongwei Wang}

\author[4]{Kecheng Liu}
\address[4]{Informatics Research Centre, Henley Business School, University of Reading, Reading, Berkshire RG6 6UD, United Kingdom}

\author[5]{Xu Wang}
\address[5]{School of Mechanical Engineering, Chongqing University, 400044, Chongqing, China}

\cortext[cor1]{Corresponding author}

\begin{abstract}
To explore the robustness of recommender systems, researchers have proposed various shilling attack models and analyzed their adverse effects. Primitive attacks are highly feasible but less effective due to simplistic handcrafted rules, while upgraded attacks are more powerful but costly and difficult to deploy because they require more knowledge from recommendations. In this paper, we explore a novel shilling attack called Graph cOnvolution-based generative shilling ATtack (GOAT) to balance the attacks' feasibility and effectiveness. GOAT adopts the primitive attacks' paradigm that assigns items for fake users by sampling and the upgraded attacks' paradigm that generates fake ratings by a deep learning-based model. It deploys a generative adversarial network (GAN) that learns the real rating distribution to generate fake ratings. Additionally, the generator combines a tailored graph convolution structure that leverages the correlations between co-rated items to smoothen the fake ratings and enhance their authenticity. The extensive experiments on two public datasets evaluate GOAT's performance from multiple perspectives. Our study of the GOAT demonstrates technical feasibility for building a more powerful and intelligent attack model with a much-reduced cost, enables analysis the threat of such an attack and guides for investigating necessary prevention measures.
\end{abstract}


\maketitle

\section{Introduction}
Recommender systems (RSs) are now widely employed in various web services such as online shopping, video-on-demand, and even web APIs \cite{Qi2020a} to increase sales or improve user experience. The dominant paradigm of RSs is collaborative filtering (CF) that suggests relevant items to users according to the collective historical user behaviors. A general principle of the CF is that similar users tend to interact with similar items. Although this principle leads to a good recommendation performance \cite{He2018}\cite{He2017}\cite{Wang2019}\cite{Qi2020b}, it also puts RSs at the risk of being manipulated. It has been indicated that recommender systems are vulnerable to malicious profile injections, which are also known as the shilling attacks \cite{Gunes2014}. With a large number of skewed ratings, injected fake user profiles can mislead the recommender system in measuring similarities \cite{Hou2018}\cite{XuePan2020}. As a result, attackers can manipulate the recommendation results for their intended commercial interests. Typically, some profit-driven raters (e.g. the item providers) may inject positive ratings to promote the reputation of their products \cite{Xia2015} and induce real users to modify their natural behavior.

To explore the robustness of recommender systems, over the past years, many studies have paid attention to the shilling attacks \cite{Gunes2014}. The early exploration of shilling attacks mainly focused on primitive attack models \cite{Lam2004}\cite{Li2016}\cite{yu2017hybrid} that craft fake user profiles based on heuristic rules. Generally, they sample items arbitrarily, and rate them with ratings subject to a specific distribution rule. Due to the simple assumption of rating distribution, these attacks are easy to deploy but less capable of causing damage. To guard against more powerful attack models, the follow-up studies \cite{Christakopoulou2019}\cite{Fang2018}\cite{lin2020attacking}\cite{tang2020revisiting} have investigated some upgraded shilling attack models that construct fake user profiles using more sophisticated methods (e.g. end-to-end training). By accessing more knowledge from recommendations, these attack models are more difficult to adderss, and hence they pose a significant threat to the security of RS. However, on the other hand, the requirement of the knowledge (e.g. access to model parameters or gradients during optimization) also give rise to a high cost involved in launching such an attack, making these upgraded attack models less practical and also less likely to be the choice of the potential adversaries.

How the attack model can reach the optimal trade-off between effectiveness and cost now becomes the primary concern of the attackers as well as of the studie on shilling attacks. To the best of our knowledge, most existing attack models struggle with this task. In most attacks' settings \cite{Lam2004}\cite{Chen2019}\cite{Pang2018}, the number of fake user ratings is usually a fixed volume, but it can be adjusted to a large value. In such a scenario, producing ratings on too many items will incur a high cost because most RS applications require users to purchase an item before rating it. Additionally, some upgraded attacks require the prior knowledge from RSs, e.g., Fang et al. \cite{Fang2018} aim to attack a graph-based RS, Chen et al. \cite{Chen2019} measure the cosine association among users, and Christakopoulou et al. \cite{Christakopoulou2019} assumes that the recommendation results are calculated by the inner product of user and item embeddings. They normally consider the generation of fake user profiles as an optimization problem, limiting the feasibility and practical use of the models. The computational cost of acquiring prior knowledge and designing of heuristic optimization algorithms is still a heavy burden for attackers. Another essential problem is that how the attack models can successfully fool the detectors against cheating. Due to the naive assumption of the rating distribution of real users or the simple choice of items for camouflage, many existing attack models can be easily detected with statistical features \cite{Li2016a}. They forge fake ratings by only considering the single or mixed rating distributions but neglect the correlation between co-rated items. That is, few of them consider rating smoothness. As a result, these strategies can lead to extreme ratings, i.e., the fake users tend to rate items with either the maximum or minimum ratings, which is not conducive to the stealth of attack.

Recently, the boom in deep learning has empowered recommendation models to achieve greater capacity, and the new techniques also provide attackers with an opportunity to create more powerful and less detectable attack models. It has been widely reported and discussed that instances with small, intentional feature perturbations can cause a deep model to make incorrect predictions that are known as adversarial example attacks \cite{yuan2019adversarial}. Although most of current adversarial attacks do not aim at recommender systems, they have provided an example showing how to imperceptibly beat machine learning models with a low cost. Therefore, it is resonable to think that analogous attacks aimed at recommender systems are also looming. To be alert for the potential emerging attacks, in this work, we investigate the possible form of novel attacks and present a deep learning-based shilling attack model called the Graph cOnvolution-based generative ATtack model (\textbf{GOAT}). The novelty of GOAT can be characterized by four main distinctions from the existing models: (i) GOAT uses a well-designed generative adversarial network (GAN) that generates fake user profiles with variable number of ratings to assure more authentic ratings. (ii) Unlike previous GAN-based approaches that generate entire item ratings, GOAT adopts a sampling method to assign rated items for fake users and only focuses on generating ratings for sampled items. (iii) For the smoothness and camouflage of fake ratings, GOAT combines a tailored graph convolution structure into the generator of GAN to capture the correlations between co-rated items and avoid extreme ratings. (iv) Rather than discovering the knowledge from recommendations, GOAT only requires the knowledge of partial user historical behaviors.

In our attack scenario, an attacker's goal is to promote target items in RSs, i.e., increase their frequency of being recommended. We note that an attacker could also degrade target items since degradation can be viewed as a special case of promotion\cite{Fang2018}. By exposing and sharing the principles and basic mechanism, we hope to make the industrial communities more ready for emerging shilling attacks resembling GOAT. In summary, our contributions are as follows.
\begin{enumerate}[\textbullet]
\item We identify the potential security vulnerability of CF-based recommender systems by presenting a new type of shilling attack. We formulate shilling attacks as a generative model that takes advantage of both handcrafted rules and end-to-end training to construct fake user profiles with less computational cost.
\item We developed a unified model GOAT that leverages GAN to enhance the authenticity of fake user profiles and integrate a graph convolution structure into GAN's generator to smoothen the fake ratings. Additionally, a sampling method is also introduced to improve the model's efficiency and effectiveness.
\item We conduct extensive experiments on two public datasets, four CF-based recommender algorithms, and three shilling attack detectors to illustrate that GOAT is more effective than other baseline models. Furthermore, we provide some suggestions for preventing and detecting such an attack by analyzing the attack effect from a multi-stakeholder perspective.
\end{enumerate}

The rest of the paper is organized as follows: In Section 2, we review the related studies. Section 3 introduces the details of the proposed shilling attack model and Section 4 gives the parameter settings of our attack model. In Section 5, our experimental results on two public datasets and analysis for preventing and detecting such an attack are presented. Finally, Section 6 concludes the paper.


\section{Related Work}
In this section, we present a brief review of the three related research directions: shilling attack, generative adversarial networks, and graph neural networks that are the important building blocks of our work.
\subsection{Shilling Attacks}
\begin{figure}[h]
\centering
\includegraphics{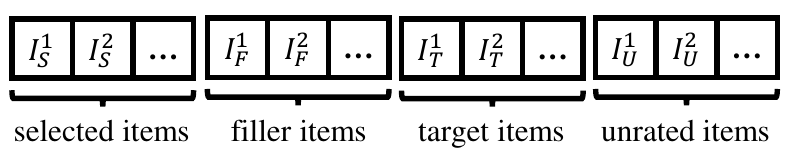}
\caption{\textrm{An illustration of a fake user profile}}
\label{Figure1}
\end{figure}

Shilling attack is a fraudulent practice that poisons recommender systems by injecting a number of fake user profiles \cite{Gunes2014}. A fake user profile usually consists of four parts -- selected items, filler items, target items, and unrated items \cite{Chen2019}; the attacker determines the size of each part, as shown in Figure \ref{Figure1}, where each space corresponds to a rated item. The selected items are those carefully selected by attackers to increase the fake user's influence; in other words, to make the fake user profiles associate with as many normal users as possible in similarity measurement. Filler items are usually used to disguise fake user profiles as the real users. Target items are a set of products that attackers aim at promoting or nuking. Since it is not practical for a user to rate all items, the rest are unrated. The primitive item selection strategy is relatively simple; attackers only consider factors such as item popularity and rating distribution \cite{Mobasher2007} \cite{Pang2018} or even adopt random strategies \cite{Lam2004}. For example, bandwagon attack \cite{Mobasher2007} chooses a certain number of most popular items as the selected items, making the generated fake user profile share overlap with a large number of real users, and then the fake user is very likely to be the nearest neighbor of these real users, resulting in a distorted recommendation list for these users. It should be noted that, some attack models such as the average attack, random attack \cite{Lam2004}, and attack proposed in \cite{Li2016} do not differentiate the selected items and filler items, which means that the same rule applies to all rated items in the fake user profiles. Likewise, some optimization-based shilling attack models use the uniform optimization algorithm to generate fake ratings. Fang et al. \cite{Fang2018} design an optimized poisoning attack for graph-based recommender systems. Chen et al. \cite{Chen2019} apply a greedy algorithm to iteratively add an item from the candidate set for item selection. Fang et al. \cite{Fang2020} leverage the influence function of a subset of influential users. Specifically, PoisonRec \cite{Song2020} is decoupled from the interactive RS under a model-free reinforcement learning architecture, but the training gradient still needs to be elaborately designed. Moreover, other variants of optimization-based models assign rated items by models rather than predefined by attackers. A deep learning-based attack proposed in \cite{Christakopoulou2019} uses GAN to generate entire user profiles and further optimizes them by approximate gradient descent. Tang et al. \cite{tang2020revisiting} refine the gradient exactness of their optimization algorithm and propose a model that applies to most embedding-based recommendations. To better control the fake rating number, we follow the four-part mechanism to construct fake user profiles in this work.

\subsection{Generative Adversarial Networks}
GAN was first proposed by Goodfellow et al. \cite{Goodfellow2014} and has led to revolutionary breakthroughs in the area of image generation. The basic idea of vanilla GAN is to alternatively optimize two neural networks that act as the discriminator and generator, by playing a minimax game. The generator is optimized to learn the distribution of the real data and fool the discriminator by using the generated data, while the discriminator is trained to distinguish the generated data from real data. Increasing research efforts on GANs have been proposed to enhance their capacity by modifying the network's architecture \cite{Zhao2017}\cite{Radford2016}, adding regularization terms to the original loss function \cite{Gulrajani2017}\cite{Liu2020} or changing the data sampling method \cite{Miyato2018}\cite{He2019}. Recently, there are also many applications of GAN on the recommendation domain \cite{Gao2021}. IRGAN \cite{Wang2017} proposes a minimax framework to unify the generative and discriminative information retrieval models. CFGAN \cite{Chae2018} proposes a GAN-based collaborative filtering framework to provide higher recommendation accuracy. RSGAN \cite{Yu2019} and ESRF \cite{Yu2020} focus on identifying reliable relations with adversarial training to improve social recommendations. Notably, recent studies on shilling attacks \cite{Christakopoulou2019}\cite{lin2020attacking} show the tendency of combining GANs with attacks. They uniformly adopt the framework of GAN to generate adversarial user profiles, such as DCGAN \cite{Radford2016} and Wasserstein GAN \cite{Gulrajani2017} are used in these works. In this work, we follow the paradigm of previous studies and design a novel GAN model to generate fake user profiles. 

\subsection{Graph Neural Networks}
The concept of GNN was first proposed by \cite{Gori2005} to learn correlations among graph data and is further elaborated in \cite{Scarselli2009}. It is a type of graph embedding technique that directly operates on the graph structure, whose principle is to capture the dependence of graphs via message passing between the nodes of graphs by using neural networks. Recently, a series of studies have demonstrated GNN's excellent performance in recommendation-related tasks. Fan et al. \cite{Fan2019} use GNN to handle social recommendations on rating prediction tasks that consider item aggregation and social aggregation for users and user aggregation for items. Wu et al. \cite{Wu2019} develop dual graph attention networks to learn representations for twofold social effects collaboratively. Moreover, a variant of GNN -- graph convolutional network (GCN) that focuses on aggregating node features from local graph neighborhoods, shows greater potential in the area of RSs. Ying et al. \cite{Ying2018} develop a data-efficient GCN algorithm for web-scale recommender systems. Wang et al. \cite{Wang2019} follow the standard GCN to refine user (or item) embeddings by aggregating the embeddings of its interacted item (or user). LightGCN \cite{He2020} argues that a standard GCN may be burdensome for recommendation and proposes a simplified model including only the most essential components in GCN. These applications also inspire us to model the high-order relationships between co-rated items via adopting GNN's mechanism. Specifically, the architecture of GCN is used in our model, and we only keep the graph convolution structure for model efficiency.

\begin{figure*}[t]
\centering
\includegraphics{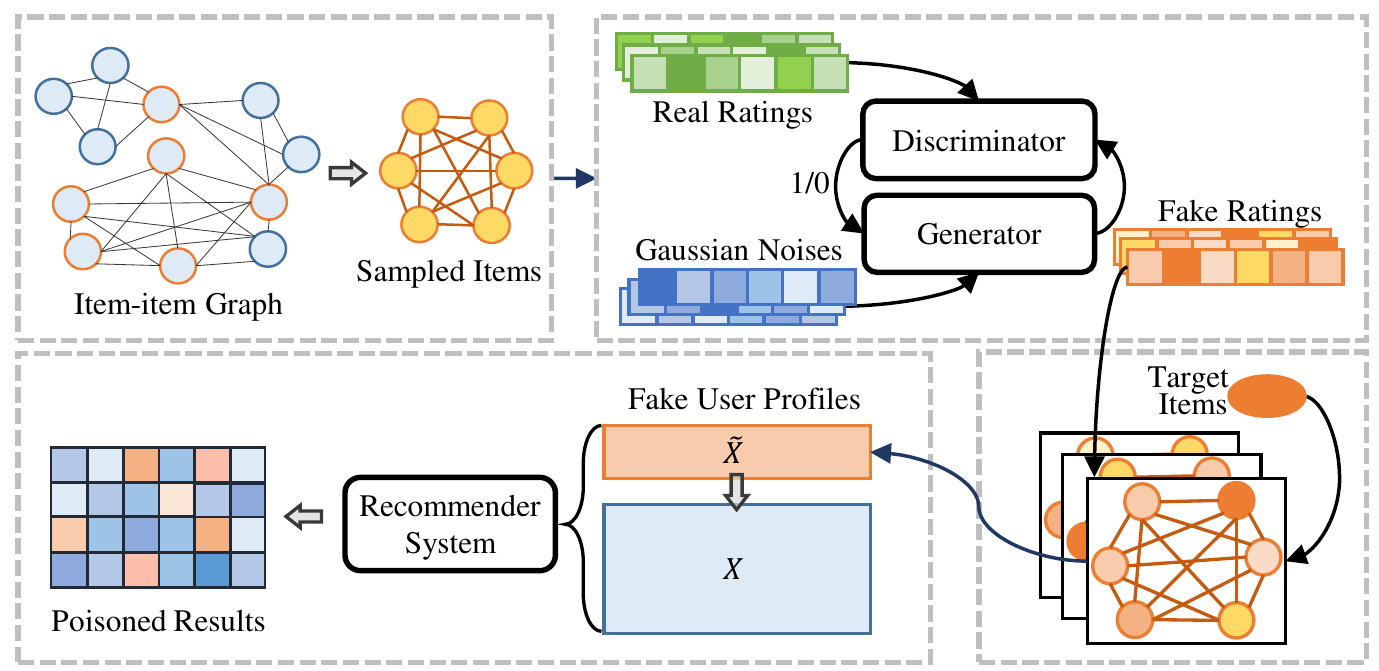}
\caption{\textrm{The framework of GOAT}}
\label{Figure2}
\end{figure*}

\section{Proposed Shilling Attack Model: GOAT}
The framework of the proposed shilling attack model -- GOAT is shown in Figure \ref{Figure2}. The pipeline mainly consists of four parts. First, items are sampled from an item-item graph to decide which items should be rated in a fake user profile. Then an adversarial architecture powered by graph convolution is trained to generate ratings for these sampled items. Next, the faker user profile is assembled by combining the sampled items and the target items (assigned high ratings for promotion). Finally, repeating above process to collect a bunch of fake user profiles and injecting them into real data.

\subsection{Sampling Items for Fake Users}
The first step of fake user generation is to determine which items will be the constituent parts. Since the users' rating behaviors reflect their preference, it is crucial to simulate the correlation of rated items from real user profiles so that fake users can influence the prediction of real users' appetites. Therefore, an item-item graph based on user-item interactions is constructed and used to sample co-rated items to construct fake user profiles for the attack model.

\textbf{Construction of the item-item graph.} The item-item graph is built on user-item interactions, as shown in Figure 3. Each interaction between users ($U$) and items ($I$) represents that a user has rated an item. This bipartite graph is converted into an item-item graph that records the co-rated relationships between items, in which each link indicates that they are jointly rated by at least one user.

\textbf{Sampling items from an item-item graph.} The selected items and filler items of each fake user are sampled from an item-item graph in two steps. A real user profile is chosen as the template, the fake rating number is determined in step (1), and items are sampled from the template user profile in step (2). The detailed steps are described as follows:
\begin{itemize}
\item [(1)] Randomly sampling a real user profile $\boldsymbol{u_i}$ that contains at least $o_u$ ratings, where $o_u$ is a threshold that ensures that the fake users will not develop from cold-start users who rated fewer than $o_u$ items. Let the number of fake ratings $k=min(|\boldsymbol{u_i}|,o_g)$, where $o_g$ is a threshold for the generator that restricts that the fake rating number should be $o_g$ at most. Then, the number of both selected items $I_S$ and filler items $I_F$ in each fake user profile is confined to $[o_u,o_g]$, and $|I_S|=k\times p_S$ and $|I_F|=k\times(1-p_S)$, where $p_S$ is the proportion of selected items.
\item [(2)] Randomly sampling items in $\boldsymbol{u_i}$ according to the item threshold $o_i$. Since the items with few ratings have little impact on user similarity measuring, we determine that the candidates of selected items should be rated at least $o_i$ times, and that the candidates of filler items should be rated at least $o_i/3$ times. If $\boldsymbol{u_i}$ does not contain adequate candidates to fill $I_S$ and/or $I_F$, we select other eligible candidates from the whole item-item graph as supplements (items connected to the current items will be given the priority).
\end{itemize}

After the above steps, the set $I_{fake}=I_S\cup{I_F}$ with items that need to be rated for a fake user is finalized.

\begin{figure}[!htbp]
\centering
\includegraphics{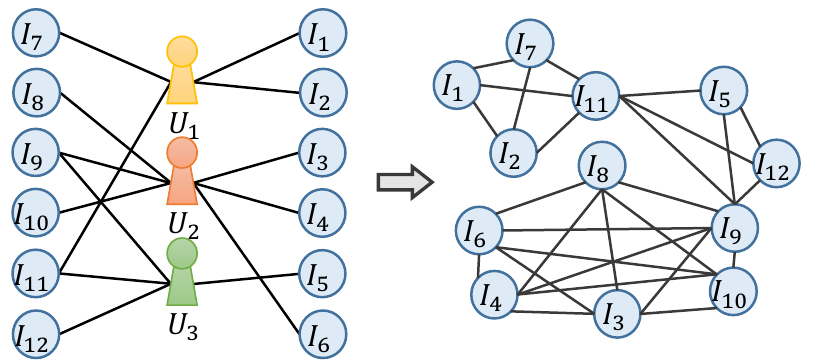}
\caption{\textrm{Converting a user-item graph to an item-item graph}}
\label{Figure3}
\end{figure}

\begin{figure}
\centering
\includegraphics{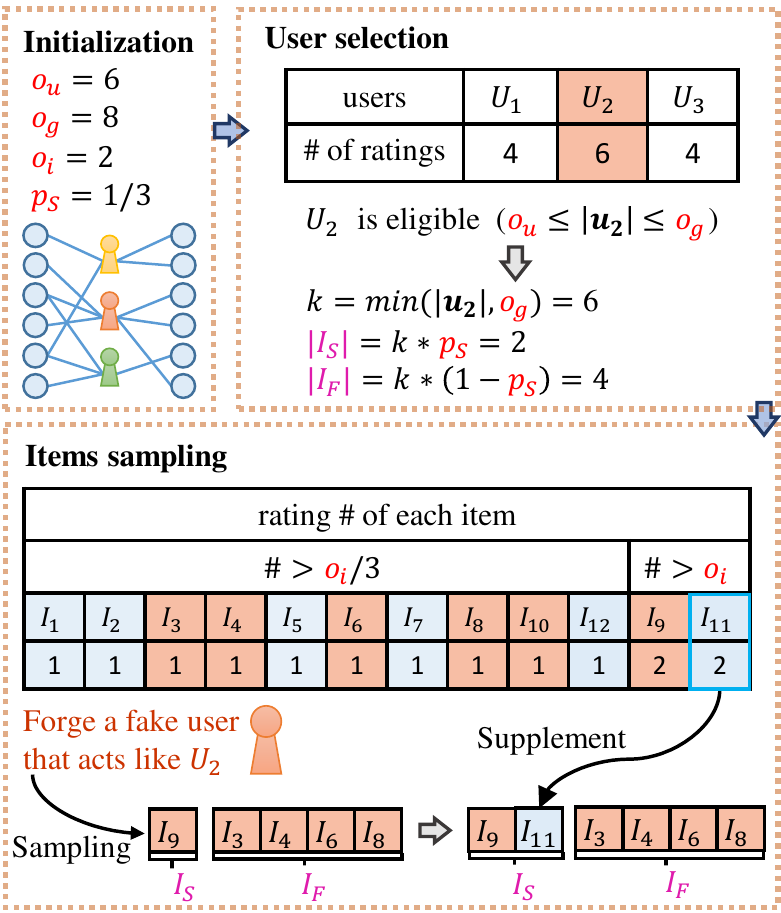}
\caption{\textrm{An illustration of sampling items for a fake user}}
\label{Figure4}
\end{figure}

\textit{An illustration:} To help gain a complete understanding of the sampling process, we provide an instance shown in Figure \ref{Figure4} that is based on the item-item graph in Figure \ref{Figure3} for illustration. Initially, we set $o_u=6$, $o_g=8$, $o_i=2$, and $p_S=1/3$ (the details of the settings will be further discussed in Section 4). Then, after step (1), only $U_2$ who rates six items is eligible. Hence, the model will forge a fake user that acts like $U_2$, and according to initial settings, we have $k=6$, $|I_S|=2$, $|I_F|=4$. Next, a fake user profile can be constructed by referring to $U_2 $'s profile. Two selected items are required, but only $I_9$ is suitable. Therefore, another selected item $I_{11}$ is sampled from the whole graph. Four filler items are required; they are randomly sampled from five candidates ($I_3, I_4, I_6, I_8, I_{10}$). Finally, the result is achieved as the set $I_{fake}=\{I_9,I_{11},I_3,I_4,I_6,I_8\}$.

\subsection{Generating Ratings for Sampled items}
After assembling the fake user profile, we show how to rate the sampled items in this section. Unlike most previous works using handcrafted ratings, in this work, a GAN powered by a graph convolution architecture is adopted to learn the rating distribution of real users. The details about generator $G$ and discriminator $D$ of our model are given in the following paragraphs.

\begin{figure}[h]
\centering
\includegraphics{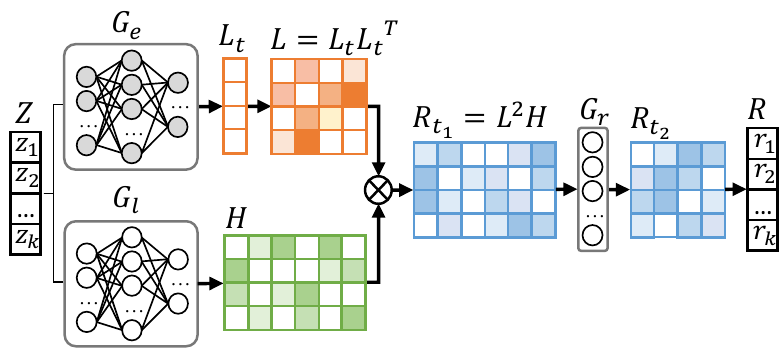}
\caption{\textrm{The generator $G$ of GOAT}}
\label{Figure5}
\end{figure}

\textbf{The Generator Model.} The overview of $G$ is depicted in Figure \ref{Figure5}. It is designed as a graph convolution structure to generate ratings for items in $I_{fake}$. The process is as follows: First, sampling $k$ noises from a Gaussian distribution as input (denoted by $Z$), where $k$ is exactly the rating number of a fake user. Second, $Z$ will be converted into rating embeddings $H$ and link representations $L$ through $G_{e}$ and $G_l$, respectively. $G_e$ is a three-layer network whose output $H$ is assumed to contain generated rating information. $G_l$ is a three-layer network whose output $L_t$ is further converted into $L=L_t{L_t}^T$ to represent the generated link weights between items. Each row of $H$ corresponds to an item's rating, and each row of $L$ contains link weights between the current item and other items. Third, the aggregation between the co-rated items is considered a weighted summation to get convolutional rating embeddings $R_{t_1}$. Each row of $R_{t_1}$ is an item's rating embedding that aggregates its neighbors' generated ratings. We consider second-order neighbors at most, which indicates that the neighboring items are not only directly co-rated by the same user but also indirectly co-rated by two users who have an intersection in their preference. Finally, a single-layer network $G_r$ is used to convert $R_{t_1}$ into $R_{t_2}$ and the model takes the average pooling $R$ on each row of $R_{t_2}$ as generated fake ratings.

\begin{figure}[!htbp]
\centering
\includegraphics{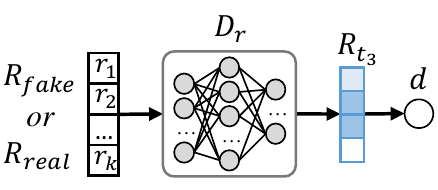}
\caption{\textrm{The discriminator $D$ of GOAT}}
\label{Figure6}
\end{figure}

\textbf{The Discriminator Model.} The discriminator model $D$ is depicted in Figure \ref{Figure6} and discriminates real user profiles from fake profiles and reinforces $G$ to generate more realistic user profiles and ratings. $D$ consists of only a simple four-layer network $D_r$. The input of $D$ is a rating vector of the item in $I_{fake}$. $D_r$ will convert the input into $R_{t_3}$, and each row contains a value used to judge whether the corresponding rating is real or fake. It considers each item's average rating in the dataset as real and the generated rating as fake. Afterward, the model takes the mean of $R_{t_3}$ to obtain the final result $d$, where the $d$ value of a real user profile should be as close to 1 as possible.

\textbf{The Final Formulation.} We follow the setting of wGAN-GP \cite{Gulrajani2017} to optimize the model because the classical cross entropy-based loss function usually leads to the collapse of optimization. The loss function of wGAN-GP is given below:
\begin{equation}
\begin{aligned}
loss&=\mathbb{E}_{z\sim N(0,1)}[D(G(z))]-\mathbb{E}_{X\sim P_{data}}[D(X)]\\
&+\lambda{[ \Vert\nabla_{\hat{X}}D(\hat{X})\Vert_2-1]^2}
\end{aligned}
\end{equation}
where $Z$ is the noise sampled from $N(0,1)$, $X$ is the ground-truth value, and $\hat{X}=\epsilon X+(1-\epsilon)G(Z), \epsilon\sim U(0,1)$. In addition, the numbers of hidden units are 128, 256, 64 in $G_e$, 128, 256, 32 in $G_l$, 128 in $G_r$, and 1024, 512, 256, 1 in $D_r$. Except for the first three layers of $D_r$ that use the sigmoid activation function, all activation functions in our model are $leaky ReLU$ with $\alpha =0.2$.
 
Additionally, inspired by gradient penalty, we add an extra regularization term as rating penalty while training $G$. Rather than clipping the abnormal ratings into a normal range, the rating penalty forces $G$ to generate normal ratings while cheating $D$. More specifically, we use $loss_D$ and $loss_G$ to optimize $D$ and $G$:
\begin{align}
loss_D&=D(G(Z))-D(X)+\lambda{[ \Vert\nabla_{\hat{X}}D(\hat{X})\Vert_2-1]^2}& \\
loss_G&=-D(G(Z))+\psi{[\frac{1}{k}\Vert{G(Z)-X}\Vert^2]}
\end{align}
where $\lambda$ and $\psi$ are hyperparameters for the gradient penalty and rating penalty, respectively. Then the model can be optimized by gradient descent until the loss functions reach their local minima simultaneously.

\begin{figure}[h]
\centering
\includegraphics{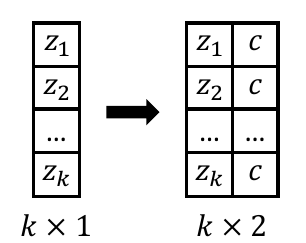}
\caption{\textrm{Adding extra information to the original input $Z$}}
\label{Figure7}
\end{figure}

\textbf{Training details.} i) Due to the variable rating number, the model must generate fake user profiles with different rating numbers. We find that a dramatic variation in the rating number causes instability in the training process. Hence, the whole training is divided into three periods to reduce the intensity of the variation. At each period, let $o_g$ as $o_g*0.5, o_g*0.7$ and $o_g$, which will be rewritten as $o_g^{'}$ in the following statement. For instance, if the training epoch is 100 and $o_g=18$, then in epochs $1-50$, we have $o_g^{'}=9$ and the generator will rate for $6\sim 9$ items; in epochs $51-70$ we have $o_g^{'}=12$ and the generator will rate for $6\sim 12$ items; and so on. ii) We find that the model has difficulty converging on a dataset with very sparse user-item interactions. The sparsity makes the generator fail to learn the rating distribution. Hence, inspired by conditional GAN \cite{Mirza2014}, extra information is added to the input $Z$, as shown in Figure \ref{Figure7}, where $c=k/o_g$ signifies the 'vitality' of a user. The more items it contains, the more significant it is. Consequently, the first layer of $G_e$ and $G_l$ in the generator is modified to $2\times 128$.

\subsection{Relation to Handcrafted Models}
The relation between handcrafted shilling attack models and GOAT can be generalized from the above two steps: sampling and rating. We give detailed illustrations in the following.

\begin{figure}[h]
\centering
\includegraphics{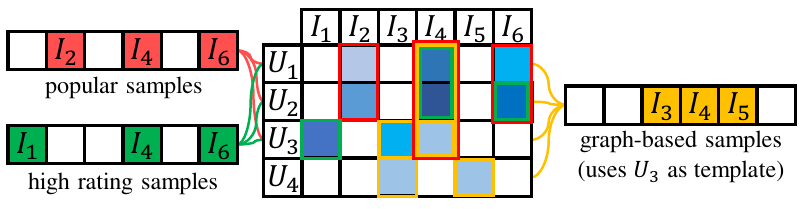}
\caption{\textrm{A toy example of sampling}}
\label{Figure8}
\end{figure}

\textbf{Sampling.} A toy model is offered in Figure \ref{Figure8} to show the potential of graph-based sampling where each colored pane is a user-item interaction (the darker color means the higher rating), and blank panes mean null interactions. Handcrafted models conduct sampling mainly based on statistical features such as rating numbers (the red box) and rating scores (the green box), limiting the correlation between fake and real users. However, GOAT uses extra item-item graphs (the yellow box) to build more correlations between fake and real users by using co-rated items. Such a sampling method extends the statistical features into a graph feature.

\textbf{Rating.} Handcrafted ratings usually sample rating scores from an assigned Gaussian distribution, i.e., $r \sim N(\mu,\sigma ^{2})$, where $\mu$ and $\sigma^2$ are the mean and variance acquired from the real rating distribution and $r$ is the sampled rating. Let $z\sim N(0,1)$ and $F(z)=\mu+z*\sigma=r$, then handcrafted models' rating formulation can be deemed as to obtain parameters $\mu$ and $\sigma$ of the function $F$. Likewise, GOAT substitutes a function $G$ for $F$, where $G$ is the generator function. Then the rating formulation is converted to optimize parameters $\theta^{*}_G$ of the generator, where $G(z)=r$. Compared with $F$, $G$ models the real data distribution in a more sophisticated manner.

In summary, GOAT uses real user profiles as the template to generate fake user profiles. The sampling and training processes are summarized in Algorithms \ref{Algorithm1} and \ref{Algorithm2}, respectively. Details about the experiments will be given in the next section.

\begin{algorithm}
\caption{Sampling Items for Fake Users}
\label{Algorithm1}
\begin{algorithmic}[1]
\Require
$o_u,o_g,o_i,p_S$;
\Ensure
items to be rated in a fake user profile;
\State Initialize $o_u,o_g,o_i,p_S$;
\For{each fake user}
\State Sample a real user profile $\boldsymbol{u_i}$ that $|\boldsymbol{u_i}|\geq o_u$;
\State Set fake rating number $k=min(|\boldsymbol{u_i}|,o_g)$;
\State Sample items under item threshold $o_i$ from $\boldsymbol{u_i}$;
\If{$\boldsymbol{u_i}$ do not contain sufficient candidate items}
\State Sample supplement items from the whole item set and those connected to the current items will be given the priority;
\EndIf
\State Combine $I_S$ and $I_F$ into set $I_{fake}$.
\EndFor
\end{algorithmic}
\end{algorithm}

\begin{algorithm}
\caption{The Optimization of the GOAT Model}
\label{Algorithm2}
\begin{algorithmic}[2]
\Require
$\theta_G,\theta_D,\lambda,\psi,\eta,Z$;
\Ensure
item ratings in a fake user profile;
\State Initialize model parameters $\theta_G$ and $\theta_D$ by Xavier initialization, set regularization parameters $\lambda, \psi$, and learning rate $\eta$;
\For{each training epoch}
\State Set the $period$ from \{0.5, 0.7, 1\}, let $o_g^{'}=o_g * period$;
\For{$D$'s training steps}
\State Sample a rating vector $X\sim{P_{data}}$ that contains $k$ ratings according to Algorithm 1;
\State Sample a noise vector $Z\sim{N(0,1)}$ that contains $k$ noises;
\State Sample $\epsilon$ from $U(0,1)$;
\State Update $\theta_D\leftarrow\theta_D-\eta *Adam(\nabla_{\theta_D}loss_D)$;
\EndFor
\For{$G$'s training steps}
\State Sample a rating vector $X\sim{P_{data}}$ that contains $k$ ratings according to Algorithm 1;
\State Sample a noise vector $Z\sim{N(0,1)}$ that contains $k$ noises;
\State Update $\theta_G\leftarrow\theta_G-\eta *Adam(\nabla_{\theta_G}loss_G)$;
\EndFor
\EndFor
\end{algorithmic}
\end{algorithm}

\section{Attack Parameter Settings}
The appropriate parameter settings of attack are vital to reduce the attack cost while maintaining the attack effect. We discuss these settings in this section.

\begin{figure}[h]
\centering
\subfigure[Douban]
{
\includegraphics{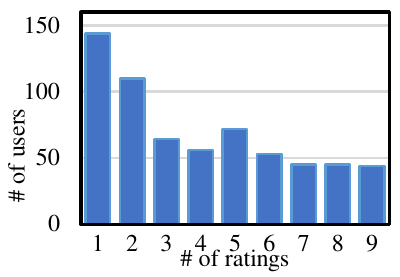}
}
\subfigure[Ciao]
{
\includegraphics{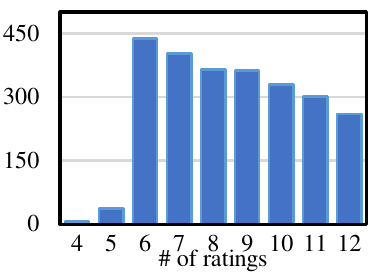}
}
\caption{\textrm{Rated item numbers among users on Douban and Ciao}}
\label{Figure9}
\end{figure}

The attacks are assumed to launch on two datasets \textbf{Douban} and \textbf{Ciao}. First, we count each user's ratings to determine the value of $o_u$ and $o_g$, confining each fake user's number of ratings to [$o_u$, $o_g$]. The statistical results are shown in Figure \ref{Figure9}, where the x-axis represents the rating number, and the y-axis represents the number of users who rate such a number of items. In our settings, we add up the values on the y-axis along the x-axis until the accumulation reaches 50\% of the total user number, and take the value at this point on the x-axis as $o_g$. That is, too many ratings are unwanted due to the need to reduce costs, so that each fake user's number of ratings just needs to cover only half of the real users' ratings at most. To this end, $o_g=35$ on Douban and $o_g=18$ on Ciao. Next, we aim to seek the most popular number of ratings to assign values for $o_u$. In other words, fake users are expected to rate enough items to ensure their attack effectiveness, so that the minimum number of ratings is selected according to real user profiles. From Figure \ref{Figure9}, most users rate no more than six items on Douban, which may be insufficient to match the maximum rating number $o_g$ (35); hence we slightly increase this value and set $o_u=8$ to make a compromise with the higher upper bound $o_g$. By contrast, the most popular number of ratings on Ciao is large enough compared to $o_g$ (18), hence $o_u=6$.

\begin{figure}[h]
\centering
\subfigure[Douban]
{
\includegraphics{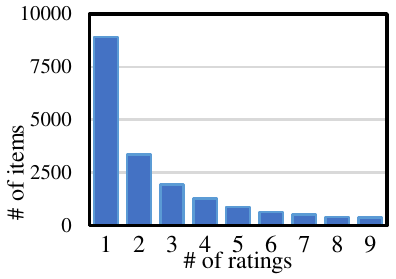}
}
\subfigure[Ciao]
{
\includegraphics{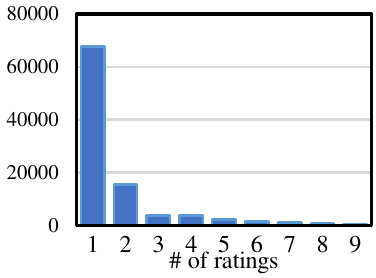}
}
\caption{\textrm{Rating numbers among items on Douban and Ciao}}
\label{Figure10}
\end{figure}

Second, we count each item's rating number to determine the value of $o_i$ that controls the candidate item selection for fake user profiles. The statistical results are shown in Figure \ref{Figure10}, where the x-axis represents the number of ratings, and the y-axis represents the number of items that are rated this number of times. For a trade-off between losing too many optional items and impairing the quality of rated items in fake user profiles, we set $o_i=8$ on both datasets. After the refinement, only 45\% items in Douban and 21\% items in Ciao are used for shilling attack model training.

Third, we set the proportion of selected items $p_S=0.3$, the learning rate $\eta=0.001$, the penalty coefficient $\lambda=10$ and $\psi=10$ on both datasets. There are 70\% of the data using for training and 30\% for testing, where the shilling attack model is trained on the training set, and the testing set is used to examine the performance of the recommender systems under attack.

In addition, since the attacking goal is to promote items, we choose items whose average ratings are smaller than '2' as target items. To avoid exposing an intense attack intent, they are rated with a subhigh rating of '4' by fake users. We assign the fake user number according to the fraction of real user numbers from 1\% to 5\%. Moreover, the trick of conditional GAN is only used on \textbf{Ciao} for its sparsity. Our model converges to the local minimum after approximately 12,000 training epochs on Douban and 15,000 on Ciao.

\section{Experiments and Analysis}
To demonstrate the effectiveness of the shilling attack model and provide guidance for the prevention and detection of such an attack, we conduct a series of experiments to analyze the attack effect from the perspective of the attackers, service providers, and detectors. The experiments focus on answering the following research questions: (RQ1) How effective are the attacks while reducing the attack cost and how do they work? (RQ2) How will the recommendation performance be affected under shilling attacks? (RQ3) Can detection methods identify the attacks? (RQ4) What prevention and detection measures are possible for counterattack?

\subsection{Experimental Designs}
\paragraph{\textbf{Dataset}}
The experiments are conducted on two public datasets: Douban and Ciao. Both datasets employ a 5-star rating system that provides integer ratings.
\begin{enumerate}[\textbullet]
\item\textbf{Douban}\footnote{http://smiles.xjtu.edu.cn/Download/download Douban.html}. The Douban dataset is collected from the Douban Movie website, a popular social media website in China. It contains 894,887 ratings given by 2,848 users to a total of 39,586 movies.
\item\textbf{Ciao}\footnote{http://www.public.asu.edu/~jtang20/datasetcode/truststudy.htm}. Ciao is a popular product review site in the UK. This dataset contains 284,086 ratings from 7,375 users on a total of 105,114 products.
\end{enumerate}

The statistics of the two datasets are listed in Table \ref{Table1}.

\begin{table*}
\centering
\caption{\textrm{The statistics of datasets}}
\begin{tabular}{|c|c|c|c|c|c|}
\hline
Dataset & \# of Users & \# of Items & \# of Ratings & Rating Scale & Sparsity(\%)\\
\hline
Douban & 2,848 & 39,586  & 894,887 & [1,5] & 99.2062\\
Ciao   & 7,375 & 105,144 & 284,086 & [1,5] & 99.9635\\
\hline
\end{tabular}
\label{Table1}
\end{table*}

\paragraph{\textbf{Baselines of Attack Models}}
The following baselines are chosen to evaluate the effectiveness of the proposed GOAT model.

\begin{enumerate}[\textbullet]
\item\textbf{Average attack}\cite{Lam2004}. In an average attack, each item's rating is compiled with a Gaussian distribution in which the mean and variance are calculated from the ratings it received. It chooses items randomly and rates each sampled item from its corresponding distribution.
\item\textbf{Random attack}\cite{Lam2004}. In a random attack, all items' ratings are compiled with a Gaussian distribution in which the mean and variance are calculated from all ratings. It chooses items randomly and rates them from the total rating distribution.
\item\textbf{Bandwagon attack}\cite{Mobasher2007}. In a bandwagon attack, the selected items are popular items, and they are rated with the highest ratings. The filler items share the same setting as those in the average attack.
\item\textbf{Unorganized malicious attack}\cite{Pang2018}. In an unorganized malicious attack, the attackers are supposed to be unorganized, i.e., it combines the aforementioned attack scheme.
\item\textbf{AppGrad attack}\cite{Christakopoulou2019}. AppGrad first initializes entire fake user profiles via a GAN trained on the user-item matrix and then achieves the attack goal by using approximate gradients to optimize fake user profiles. Since AppGrad generates ratings for entire items, we perform downsampling to cut the number of fake ratings to fit our attack cost setting.
\item\textbf{Augmented shilling attack}\cite{lin2020attacking}.The augmented shilling attack is implemented with GAN. It generates fake user profiles by augmenting the "template" of real user profiles; i.e., it directly samples the filler items' ratings from real user profiles as input to generate fake ratings for selected items and then combines both parts as fake user profiles.
\item\textbf{Influence-function-based attack}\cite{Fang2020}. The influence-function-based attack leverages the influence function to measure each data point's influence on the recommending target items, and generates fake user profiles by optimizing the attack goal on a set of real users with the largest influence.
\end{enumerate}

These shilling attack models are summarized in Table \ref{Table2}, and for simplicity, we abbreviate these attack models and our attack models as Ave, Ran, Band, UM, AG, AUSH, IF and GOAT, respectively.

\begin{table*}
\centering
\caption{\textrm{Comparison between different shilling attack models}}
\begin{tabular}{|c|c|c|c|c|c|}
\hline
& \multicolumn{4}{c|}{Composition of fake user profile} & \\
\cline{2-5}
\multirow{-2}{*}{\begin{tabular}[c]{@{}c@{}}Attack \\ model\end{tabular}} & Selected items & Filler items & \begin{tabular}[c]{@{}c@{}}Target\\ items\end{tabular} & \begin{tabular}[c]{@{}c@{}}Unrated\\ items\end{tabular} & \multirow{-2}{*}{Required knowledge} \\
\midrule[1.5pt]
Ave & \multicolumn{2}{c|}{\begin{tabular}[c]{@{}c@{}} Random items with ratings from \\ their corresponding Gaussian distributions \end{tabular}} & & & \begin{tabular}[c]{@{}c@{}} Mean and variance on \\ ratings of each item\end{tabular} \\
\cline{1-3}
\cline{6-6}
Ran & \multicolumn{2}{c|}{\begin{tabular}[c]{@{}c@{}} Random items with ratings from \\ a Gaussian distribution \end{tabular}} & & & \begin{tabular}[c]{@{}c@{}} Mean and variance on\\ entire ratings \end{tabular} \\
\cline{1-3}
\cline{6-6}
Band & \begin{tabular}[c]{@{}c@{}}Popular items \\ with highest ratings\end{tabular} & \begin{tabular}[c]{@{}c@{}}Random items with ratings \\ from their corresponding \\ Gaussian distribution\end{tabular} & & & \begin{tabular}[c]{@{}c@{}}Mean and variance on \\ ratings of each item;\\ item popularity\end{tabular} \\
\cline{1-3}
\cline{6-6}
UM & \multicolumn{2}{c|}{\begin{tabular}[c]{@{}c@{}} A hybrid strategy that \\ combines several attacks \end{tabular}} & \multirow{-4}{*}{\begin{tabular}[c]{@{}c@{}}Low \\ rating \\ items \\ with \\ sub- \\ high \\ ratings \end{tabular}} & \multirow{-3}{*}{\begin{tabular}[c]{@{}c@{}} Zero\\ rating \end{tabular}} & \begin{tabular}[c]{@{}c@{}} Mean and variance on \\ entire ratings and \\ ratings of each item;\\ item popularity\end{tabular} \\
\cline{1-3}
\cline{6-6}
GOAT & \multicolumn{2}{c|}{\begin{tabular}[c]{@{}c@{}}Items refined by $o_i$ with ratings\\ generated by the generator\end{tabular}} & & & \begin{tabular}[c]{@{}c@{}}Mean rating of each item;\\ users' behaviors on items\end{tabular} \\
\cline{1-3}
\cline{6-6}
AUSH & \begin{tabular}[c]{@{}c@{}}Human-selected items\\ with generated ratings\end{tabular} & \begin{tabular}[c]{@{}c@{}}Sampled items with ratings \\ from real user profiles\end{tabular} & & &  \\
\cline{1-3}
IF & \multicolumn{2}{c|}{\begin{tabular}[c]{@{}c@{}}Items that have the largest optimized values with \\ ratings from their corresponding Gaussian distributions\end{tabular}} & & &  \multirow{-2}{*}{\begin{tabular}[c]{@{}c@{}}User ratings and their\\ behaviors on items\end{tabular}} \\
\cline{1-5}
AG & \multicolumn{4}{c|}{\begin{tabular}[c]{@{}c@{}} Entire items with optimized ratings\end{tabular}} & \\
\hline
\end{tabular}
\label{Table2}
\end{table*}

\paragraph{\textbf{Target Recommendation Algorithms}} The attacks are deployed on four popular RSs:

\begin{enumerate}[\textbullet]
\item\textbf{BPR}\cite{Rendle2009}. Bayesian personalized ranking is a generic optimization criterion that maximizes the posterior probability of a Bayesian analysis problem. It learns a matrix factorization model by the pairwise ranking objective function.
\item\textbf{APR}\cite{He2018}. Adversarial personalized ranking performs adversarial training to enhance the pairwise ranking method BPR. It adds adversarial perturbations on parameters while optimizing the model.
\item\textbf{NeuMF}\cite{He2017}. Neural matrix factorization unifies the linearity of matrix factorization and non-linearity of the multilayer perceptron to model the collaborative filtering structures.
\item\textbf{NGCF}\cite{Wang2019}. Neural graph collaborative filtering exploits the user-item graph by propagating user embeddings and item embeddings, leading to the expressive modeling of high-order connectivity in the user-item graph.
\end{enumerate}

\paragraph{\textbf{Shilling Attack Detection Algorithms}} 
A principal component analysis (PCA)-based \cite{Mehta2009}, expectation maximization (EM)-based \cite{Cao2013}, and item popularity-based \cite{Li2016a} methods are selected as representative unsupervised, semi-supervised and supervised shilling attack detection algorithms respectively.

\paragraph{\textbf{Evaluation Metrics}}
The effectiveness of shilling attacks is evaluated by using the hit ratio ($HR$@10) \cite{Burke2015}, i.e., the ratio of target items that appear in real users' top ten recommendation lists. As given in formula (4), where $I_{target}$ is the set of target items, $U$ is the set of real users, and $\delta(u,i)$ is  an indicator function that equals 1 if item $i$ hits $u$'s recommendation list and $0$ otherwise.
\begin{equation}
\begin{aligned}
HR@10=\frac{1}{|I_{target}|}\sum_{i\in{I_{target}}}\frac{1}{|U|}\sum_{u\in{U}}\delta{(u,i)}
\end{aligned}
\end{equation}

$Precision$ and $NDCG$(normalized discounted cumulative gain) \cite{Herlocker2004} metrics are introduced to evaluate the recommendation performance under attack. As given in formulas (5) and (6), $P$ and $T$ denote the predicted results and the real results of the top ten recommendation lists, respectively, and $\delta(P_u^j, T_u)$ is an indicator function that equals 1 if $u$'s real recommendation list contains the $j$-th recommended item and 0 otherwise.
\begin{align}
Precision@10&=\frac{1}{|U|}\sum_{u\in{U}}\frac{|P_u\cap{T_u}|}{10} \\
NDCG@10&=\frac{1}{|U|}\sum_{u\in{U}}\sum_{j=1}^{10}\frac{2^{\delta{(P_u^j, T_u)-1}}}{log_2{(j+1)}}
\end{align}

In addition, two classification metrics, $Precision$ and $Recall$ are used to evaluate the detection of shilling attacks. The definitions are given in formulas (7) and (8), where $TP$ is the number of correctly classified fake users, $FP$ is the number of misclassified real users, and $FN$ is the number of misclassified fake users.
\begin{align}
Precision&=\frac{TP}{TP+FP} \\
Recall&=\frac{TP}{TP+FN}
\end{align}

\subsection{The Attack Effects of Shilling Attack Models}
This section addresses the first research question (RQ1) regarding the attack effectiveness, cost, and mechanism of our shilling attack model.

\paragraph{\textbf{Attack Effects}}
In this experiment, we run each recommendation algorithm ten times to obtain the $HR@10$ under different injection ratios and then take the median of the results to evaluate the attack effect of each attack model (Tables \ref{Table3} and \ref{Table4})\footnote{Some results in the tables have the same data values but are not equal because of the numerical accuracy. For example, Ave and UM's results are the same on BPR when the injecting ratio is 0.01 in Table \ref{Table3} (Douban (multiple)). However, after further improving the numerical accuracy, the results are 0.0203\% and 0.0163\%, respectively.}. For the number of target items, we have two settings: `single', which promotes one specific item, and `multiple', which promotes a set of (ten) items. None of the target items hit any recommendation list prior to the attack.

\begin{table*}
\centering
\caption{\textrm{Attack effects on Douban}}
\label{Table3}
\begin{tabular}{|c|c|c|c|c|c|c|c|c|c|c|c|}
\hline
{} &
{Algorithm} &
\multicolumn{5}{c|}{{BPR}} &
\multicolumn{5}{c|}{{APR}} \\ \cline{2-12}
{} &
{Fraction} &
{0.01} &
{0.02} &
{0.03} &
{0.04} &
{0.05} &
{0.01} &
{0.02} &
{0.03} &
{0.04} &
{0.05} \\ \cline{2-12}
{} &
{Ave} &
{0.00\%} &
{0.12\%} &
{0.28\%} &
{0.61\%} &
{1.22\%} &
{0.00\%} &
{0.16\%} &
{0.24\%} &
{0.20\%} &
{0.24\%} \\ \cline{2-12}
{} &
{Ran} &
{0.00\%} &
{0.04\%} &
{0.12\%} &
{0.69\%} &
{1.02\%} &
{\textbf{0.04\%}} &
{0.12\%} &
{0.24\%} &
{0.41\%} &
\underline{0.41\%} \\ \cline{2-12}
{} &
{Band} &
{0.00\%} &
{0.00\%} &
{0.16\%} &
{0.77\%} &
\underline{0.65\%} &
{0.00\%} &
{0.16\%} &
{0.20\%} &
\underline{0.20\%} &
{0.28\%} \\ \cline{2-12}
{} &
{UM} &
{0.00\%} &
{0.04\%} &
{0.08\%} &
{0.37\%} &
\underline{0.33\%} &
{\textbf{0.04\%}} &
{0.24\%} &
{0.49\%} &
{0.57\%} &
{0.85\%} \\ \cline{2-12}
{} &
{AG} &
{0.00\%} &
\textbf{0.57\%} &
{1.30\%} &
{2.07\%} &
{2.84\%} &
\textbf{0.04\%} &
{0.16\%} &
{6.87\%} &
{8.78\%} &
{10.48\%} \\ \cline{2-12}
{} &
{AUSH} &
{0.00\%} &
{0.00\%} &
{0.16\%} &
{1.91\%} &
{17.64\%} &
{0.00\%} &
{0.00\%} &
{0.04\%} &
{0.57\%} &
{19.63\%} \\ \cline{2-12}
{} &
{IF} &
{0.00\%} &
{0.12\%} &
{1.54\%} &
{14.75\%} &
{39.54\%} &
{0.00\%} &
{0.08\%} &
{0.16\%} &
{0.57\%} &
{1.54\%} \\ \cline{2-12}
{} &
{GOAT} &
{0.00\%} &
{0.08\%} &
{\textbf{3.17\%}} &
{\textbf{30.23\%}} &
{\textbf{63.55\%}} &
{0.00\%} &
{\textbf{1.02\%}} &
{\textbf{34.17\%}} &
{\textbf{75.58\%}} &
{\textbf{88.66\%}} \\ \cline{2-12}
{} &
{Algorithm} &
\multicolumn{5}{c|}{{NeuralMF}} &
\multicolumn{5}{c|}{{NGCF}} \\ \cline{2-12}
{} &
{Fraction} &
{0.01} &
{0.02} &
{0.03} &
{0.04} &
{0.05} &
{0.01} &
{0.02} &
{0.03} &
{0.04} &
{0.05} \\ \cline{2-12}
{} &
{Ave} &
{0.00\%} &
{0.57\%} &
{0.85\%} &
{1.79\%} &
{4.10\%} &
{0.08\%} &
{0.12\%} &
{0.20\%} &
{0.16\%} &
{0.12\%} \\ \cline{2-12}
{} &
{Ran} &
{0.00\%} &
{\textbf{0.77\%}} &
{0.98\%} &
{1.54\%} &
{4.23\%} &
{\textbf{0.12\%}} &
{0.16\%} &
{0.24\%} &
{0.24\%} &
{0.12\%} \\ \cline{2-12}
{} &
{Band} &
{0.00\%} &
{0.00\%} &
{0.00\%} &
{0.04\%} &
{0.08\%} &
{\textbf{0.12\%}} &
\underline{0.08\%} &
\underline{0.08\%} &
\underline{0.00\%} &
{0.04\%} \\ \cline{2-12}
{} &
{UM} &
{0.00\%} &
{0.20\%} &
{0.41\%} &
{0.85\%} &
{2.15\%} &
{0.08\%} &
{0.24\%} &
{0.53\%} &
{0.16\%} &
{0.24\%} \\ \cline{2-12}
\multirow{-14}{*}{{\begin{tabular}[c]{@{}c@{}}Douban\\ (single)\end{tabular}}} &
{AG} &
\textbf{0.16\%} &
{0.20\%} &
{0.28\%} &
{0.65\%} &
{0.73\%} &
{0.08\%} &
{0.12\%} &
{0.16\%} &
\underline{0.12\%} &
{0.24\%} \\ \cline{2-12}
{} &
{AUSH} &
{0.00\%} &
{0.33\%} &
{1.58\%} &
{3.45\%} &
{6.70\%} &
{0.00\%} &
{0.41\%} &
{0.98\%} &
{\underline{0.85\%}} &
{2.48\%} \\ \cline{2-12}
{} &
{IF} &
{0.00\%} &
{0.37\%} &
{1.79\%} &
{3.90\%} &
{6.58\%} &
{0.00\%} &
{0.28\%} &
{0.57\%} &
{1.10\%} &
{2.72\%} \\ \cline{2-12}
{} &
{GOAT} &
{0.00\%} &
{0.37\%} &
{\textbf{2.19\%}} &
{\textbf{5.49\%}} &
{\textbf{11.17\%}} &
{0.04\%} &
{\textbf{0.49\%}} &
{\textbf{1.02\%}} &
{\textbf{1.22\%}} &
{\textbf{3.09\%}} \\ \hline
{} &
{Algorithm} &
\multicolumn{5}{c|}{{BPR}} &
\multicolumn{5}{c|}{{APR}} \\ \cline{2-12}
{} &
{Fraction} &
{0.01} &
{0.02} &
{0.03} &
{0.04} &
{0.05} &
{0.01} &
{0.02} &
{0.03} &
{0.04} &
{0.05} \\ \cline{2-12}
{} &
{Ave} &
{0.02\%} &
{0.08\%} &
{0.11\%} &
{0.12\%} &
{0.13\%} &
{0.03\%} &
{0.07\%} &
{0.09\%} &
{0.09\%} &
{0.12\%} \\ \cline{2-12}
{} &
{Ran} &
{0.00\%} &
{0.07\%} &
{0.04\%} &
{0.08\%} &
{0.19\%} &
{0.02\%} &
{0.05\%} &
{0.11\%} &
{0.17\%} &
{0.25\%} \\ \cline{2-12}
{} &
{Band} &
{0.00\%} &
{0.02\%} &
{0.06\%} &
{0.07\%} &
{0.12\%} &
{0.11\%} &
{0.08\%} &
{0.09\%} &
{0.12\%} &
{0.12\%} \\ \cline{2-12}
{} &
{UM} &
{0.02\%} &
{0.08\%} &
{0.16\%} &
{0.23\%} &
{0.26\%} &
{0.02\%} &
{0.04\%} &
{0.06\%} &
{0.12\%} &
{0.20\%} \\ \cline{2-12}
{}&
{AG} &
\textbf{0.05\%} &
{0.98\%} &
{1.34\%} &
{1.49\%} &
\underline{1.41\%} &
{0.20\%} &
{0.50\%} &
{1.07\%} &
{1.78\%} &
{2.56\%} \\ \cline{2-12}
{} &
{AUSH} &
{0.00\%} &
{\textbf{2.20\%}} &
{2.96\%} &
{4.88\%} &
{5.40\%} &
{0.12\%} &
{0.49\%} &
{1.04\%} &
{1.54\%} &
{2.65\%} \\ \cline{2-12}
{} &
{IF} &
{0.00\%} &
{1.89\%} &
{3.61\%} &
{5.38\%} &
{7.47\%} &
{0.18\%} &
{0.43\%} &
{0.89\%} &
{1.21\%} &
{2.22\%} \\ \cline{2-12}
{} &
{GOAT} &
{0.00\%} &
{2.00\%} &
{\textbf{5.86\%}} &
{\textbf{6.79\%}} &
{\textbf{7.71\%}} &
{\textbf{0.28\%}} &
{\textbf{0.61\%}} &
{\textbf{1.25\%}} &
{\textbf{1.94\%}} &
{\textbf{2.78\%}} \\ \cline{2-12}
{} &
{Algorithm} &
\multicolumn{5}{c|}{{NeuralMF}} &
\multicolumn{5}{c|}{{NGCF}} \\ \cline{2-12}
{} &
{Fraction} &
{0.01} &
{0.02} &
{0.03} &
{0.04} &
{0.05} &
{0.01} &
{0.02} &
{0.03} &
{0.04} &
{0.05} \\ \cline{2-12}
{} &
{Ave} &
{0.03\%} &
{0.51\%} &
{0.81\%} &
{1.53\%} &
{2.52\%} &
{0.00\%} &
{0.00\%} &
{0.02\%} &
{0.00\%} &
{0.00\%} \\ \cline{2-12}
{} &
{Ran} &
{0.17\%} &
{0.46\%} &
{3.05\%} &
{4.17\%} &
{5.14\%} &
{0.00\%} &
{0.02\%} &
{0.03\%} &
{0.02\%} &
{0.04\%} \\ \cline{2-12}
{} &
{Band} &
{0.07\%} &
{0.11\%} &
{0.24\%} &
{0.30\%} &
{0.35\%} &
\textbf{0.03\%} &
{0.03\%} &
\underline{0.01\%} &
{0.02\%} &
\underline{0.01\%} \\ \cline{2-12}
{} &
{UM} &
{0.03\%} &
{0.15\%} &
{1.22\%} &
{1.98\%} &
{3.07\%} &
{0.02\%} &
{0.04\%} &
{0.06\%} &
\underline{0.00\%} &
{0.08\%} \\ \cline{2-12}
\multirow{-14}{*}{{\begin{tabular}[c]{@{}c@{}}Douban\\ (multiple)\end{tabular}}} &
{AG} &
{0.09\%} &
{0.24\%} &
{0.34\%} &
{0.44\%} &
{1.32\%} &
{0.02\%} &
{0.03\%} &
{0.04\%} &
\underline{0.03\%} &
{0.05\%} \\ \cline{2-12}
{} &
{AUSH} &
{0.02\%} &
{0.53\%} &
{0.87\%} &
{1.58\%} &
{3.17\%} &
{0.01\%} &
{\textbf{0.07\%}} &
{\underline{0.06\%}} &
{0.12\%} &
{0.15\%} \\ \cline{2-12}
{}&
{IF} &
{0.12\%} &
{0.56\%} &
{0.80\%} &
{1.79\%} &
{2.72\%} &
{0.00\%} &
{0.05\%} &
{0.09\%} &
\textbf{0.13\%} &
{0.15\%} \\ \cline{2-12}
{} &
{GOAT} &
{\textbf{1.63\%}} &
{\textbf{2.32\%}} &
{\textbf{3.15\%}} &
{\textbf{5.27\%}} &
{\textbf{6.92\%}} &
{0.00\%} &
{0.02\%} &
{\textbf{0.09\%}} &
{\textbf{0.13\%}} &
{\textbf{0.17\%}} \\ \hline
\end{tabular}
\end{table*}


\begin{table*}
\centering
\caption{\textrm{Attack effects on Ciao}}
\label{Table4}
\begin{tabular}{|c|c|c|c|c|c|c|c|c|c|c|c|}
\hline
{} &
{Algorithm} &
\multicolumn{5}{c|}{{BPR}} &
\multicolumn{5}{c|}{{APR}} \\ \cline{2-12}
{} &
{Fraction} &
{0.01} &
{0.02} &
{0.03} &
{0.04} &
{0.05} &
{0.01} &
{0.02} &
{0.03} &
{0.04} &
{0.05} \\ \cline{2-12}
{} &
{Ave} &
{0.58\%} &
{1.08\%} &
{1.51\%} &
\underline{1.08\%} &
\underline{0.96\%} &
{0.54\%} &
{3.42\%} &
\underline{1.25\%} &
{3.79\%} &
{7.29\%} \\ \cline{2-12}
{} &
{Ran} &
{0.58\%} &
{0.78\%} &
{0.94\%} &
{1.19\%} &
{1.29\%} &
{\textbf{4.49\%}} &
\underline{2.01\%} &
{3.49\%} &
{4.15\%} &
{4.40\%} \\ \cline{2-12}
{} &
{Band} &
{0.60\%} &
\underline{0.60\%} &
{1.15\%} &
\underline{0.56\%} &
\underline{0.56\%} &
{0.61\%} &
{5.03\%} &
{9.95\%} &
\underline{9.10\%} &
{14.83\%} \\ \cline{2-12}
{} &
{UM} &
{0.38\%} &
{0.58\%} &
{0.69\%} &
{3.08\%} &
\textbf{6.91\%} &
{0.08\%} &
{2.49\%} &
{3.60\%} &
{7.03\%} &
{28.32\%} \\ \cline{2-12}
{} &
{AG} &
{0.43\%} &
{0.54\%} &
{0.67\%} &
\underline{0.65\%} &
{0.72\%} &
{0.64\%} &
{5.42\%} &
{10.18\%} &
{14.84\%} &
{15.67\%} \\ \cline{2-12}
{} &
{AUSH} &
{0.58\%} &
{0.96\%} &
{\textbf{2.11\%}} &
{\underline{0.93\%}} &
{4.89\%} &
{0.46\%} &
{2.43\%} &
{12.17\%} &
{21.15\%} &
{36.49\%} \\ \cline{2-12}
{} &
{IF} &
{0.51\%} &
{1.88\%} &
\underline{1.57\%} &
\textbf{3.20\%} &
{4.34\%} &
{0.61\%} &
{1.82\%} &
{3.52\%} &
{17.44\%} &
{33.60\%} \\ \cline{2-12}
{} &
{GOAT} &
{\textbf{0.68\%}} &
{\textbf{2.08\%}} &
\underline{1.63\%} &
{2.01\%} &
{3.46\%} &
{0.96\%} &
{\textbf{8.43\%}} &
{\textbf{15.94\%}} &
{\textbf{26.78\%}} &
{\textbf{47.58\%}} \\ \cline{2-12}
{} &
{Algorithm} &
\multicolumn{5}{c|}{{NeuralMF}} &
\multicolumn{5}{c|}{{NGCF}} \\ \cline{2-12}
{} &
{Fraction} &
{0.01} &
{0.02} &
{0.03} &
{0.04} &
{0.05} &
{0.01} &
{0.02} &
{0.03} &
{0.04} &
{0.05} \\ \cline{2-12}
{} &
{Ave} &
{0.00\%} &
{0.00\%} &
{0.00\%} &
{0.00\%} &
{0.00\%} &
{0.07\%} &
{0.19\%} &
{0.26\%} &
{0.29\%} &
{0.54\%} \\ \cline{2-12}
{} &
{Ran} &
{0.00\%} &
{0.00\%} &
{0.00\%} &
{0.00\%} &
{0.03\%} &
{0.10\%} &
{0.18\%} &
{0.21\%} &
{0.29\%} &
\underline{0.25\%} \\ \cline{2-12}
{} &
{Band} &
{0.00\%} &
{0.00\%} &
{0.00\%} &
{0.00\%} &
{0.00\%} &
{0.00\%} &
{0.00\%} &
{0.00\%} &
{0.17\%} &
{0.53\%} \\ \cline{2-12}
{} &
{UM} &
{0.00\%} &
{0.00\%} &
{0.00\%} &
{\textbf{0.28\%}} &
{0.39\%} &
{0.13\%} &
{0.31\%} &
{0.36\%} &
{0.81\%} &
{0.90\%} \\ \cline{2-12}
\multirow{-14}{*}{{\begin{tabular}[c]{@{}c@{}}Ciao\\ (single)\end{tabular}}} &
{AG} &
{0.00\%} &
{0.00\%} &
{0.01\%} &
{0.03\%} &
{0.04\%} &
{0.11\%} &
{0.26\%} &
{0.76\%} &
{0.97\%} &
{1.14\%} \\ \cline{2-12}
{} &
{AUSH} &
{0.00\%} &
{0.00\%} &
{0.00\%} &
{0.01\%} &
{0.03\%} &
{\textbf{0.25\%}} &
{0.32\%} &
{0.71\%} &
{0.86\%} &
{1.15\%} \\ \cline{2-12}
{} &
{IF} &
{0.00\%} &
{0.00\%} &
{0.00\%} &
{0.14\%} &
{0.38\%} &
{0.07\%} &
{0.24\%} &
{0.79\%} &
\textbf{1.35\%} &
\underline{1.24\%} \\ \cline{2-12}
{} &
{GOAT} &
{0.00\%} &
{0.00\%} &
\textbf{0.03\%} &
{0.11\%} &
\textbf{0.49\%} &
{0.11\%} &
\textbf{0.33\%} &
\textbf{0.85\%} &
{1.03\%} &
\textbf{1.47\%} \\ \hline
{} &
{Algorithm} &
\multicolumn{5}{c|}{{BPR}} &
\multicolumn{5}{c|}{{APR}} \\ \cline{2-12}
{} &
{Fraction} &
{0.01} &
{0.02} &
{0.03} &
{0.04} &
{0.05} &
{0.01} &
{0.02} &
{0.03} &
{0.04} &
{0.05} \\ \cline{2-12}
{} &
{Ave} &
{0.00\%} &
{0.00\%} &
{0.00\%} &
{0.00\%} &
{0.01\%} &
{5.05\%} &
{5.35\%} &
{7.14\%} &
{8.02\%} &
{8.27\%} \\ \cline{2-12}
{} &
{Ran} &
{0.00\%} &
{0.01\%} &
{0.01\%} &
{0.02\%} &
\underline{0.02\%} &
{\textbf{6.08\%}} &
{6.14\%} &
{6.35\%} &
{6.52\%} &
{7.05\%} \\ \cline{2-12}
{} &
{Band} &
{0.05\%} &
{0.08\%} &
{\textbf{0.11\%}} &
{0.17\%} &
{0.14\%} &
{2.54\%} &
{2.55\%} &
{2.94\%} &
{3.06\%} &
{3.52\%} \\ \cline{2-12}
{} &
{UM} &
{0.01\%} &
{0.02\%} &
{0.04\%} &
{0.05\%} &
{0.06\%} &
{2.85\%} &
{3.58\%} &
{4.68\%} &
{5.37\%} &
{5.66\%} \\ \cline{2-12}
{} &
{AG} &
{0.04\%} &
{0.05\%} &
{0.09\%} &
{0.15\%} &
{0.18\%} &
{4.10\%} &
{6.67\%} &
{7.13\%} &
{8.68\%} &
{8.71\%} \\ \cline{2-12}
{} &
{AUSH} &
{0.05\%} &
{\underline{0.02\%}} &
{0.06\%} &
{0.13\%} &
{0.13\%} &
{4.31\%} &
{7.40\%} &
{8.41\%} &
{\underline{8.19\%}} &
{9.10\%} \\ \cline{2-12}
{} &
{IF} &
{0.04\%} &
\textbf{0.09\%} &
\underline{0.08\%} &
{0.17\%} &
{0.18\%} &
{5.05\%} &
{6.91\%} &
{8.45\%} &
{8.98\%} &
{9.53\%} \\ \cline{2-12}
{} &
{GOAT} &
\textbf{0.06\%} &
{0.08\%} &
{0.10\%} &
\textbf{0.17\%} &
\textbf{0.22\%} &
{5.67\%} &
\textbf{7.64\%} &
\textbf{9.32\%} &
\textbf{9.42\%}&
\textbf{10.16\%} \\ \cline{2-12}
{} &
{Algorithm} &
\multicolumn{5}{c|}{{NeuralMF}} &
\multicolumn{5}{c|}{{NGCF}} \\ \cline{2-12}
{} &
{Fraction} &
{0.01} &
{0.02} &
{0.03} &
{0.04} &
{0.05} &
{0.01} &
{0.02} &
{0.03} &
{0.04} &
{0.05} \\ \cline{2-12}
{} &
{Ave} &
{0.00\%} &
{0.01\%} &
{0.01\%} &
{0.03\%} &
{0.09\%} &
{0.01\%} &
{0.02\%} &
{0.02\%} &
{0.02\%} &
{0.03\%} \\ \cline{2-12}
{} &
{Ran} &
{0.04\%} &
{0.31\%} &
{1.35\%} &
{3.25\%} &
\textbf{8.29\%} &
{0.02\%} &
{0.06\%} &
{0.05\%} &
{0.04\%} &
{0.05\%} \\ \cline{2-12}
{} &
{Band} &
{0.01\%} &
{0.02\%} &
{0.03\%} &
{0.05\%} &
{0.12\%} &
{0.02\%} &
{0.09\%} &
{0.12\%} &
{0.17\%} &
{0.18\%} \\ \cline{2-12}
{} &
{UM} &
{0.01\%} &
{0.16\%} &
{0.71\%} &
{2.06\%} &
{2.41\%} &
{0.01\%} &
{0.03\%} &
{0.05\%} &
{0.06\%} &
{0.07\%} \\ \cline{2-12}
\multirow{-14}{*}{{\begin{tabular}[c]{@{}c@{}}Ciao\\ (multiple)\end{tabular}}} &
{AG} &
{0.02\%} &
{0.03\%} &
{0.04\%} &
{0.05\%} &
{0.11\%} &
{0.01\%} &
{0.04\%} &
\underline{0.03\%} &
{0.05\%} &
\underline{0.05\%} \\ \cline{2-12}
{} &
{AUSH} &
{0.02\%} &
{0.32\%} &
{\textbf{1.46\%}} &
{2.62\%} &
{6.12\%} &
{0.02\%} &
{0.08\%} &
{0.09\%} &
{0.15\%} &
{0.21\%} \\ \cline{2-12}
{}&
{IF} &
{0.01\%} &
{0.24\%} &
{0.88\%} &
{2.21\%} &
{4.15\%} &
\textbf{0.13\%} &
\underline{0.11\%} &
{0.16\%} &
{0.23\%} &
{0.26\%} \\ \cline{2-12}
{} &
{GOAT} &
\textbf{0.07\%} &
\textbf{0.44\%} &
{1.16\%} &
\textbf{3.57\%} &
{4.99\%} &
{0.01\%} &
\textbf{0.13\%} &
\textbf{0.17\%} &
\textbf{0.26\%} &
\textbf{0.29\%} \\ \hline
\end{tabular}
\end{table*}

Tables \ref{Table3} and \ref{Table4} indicate that GOAT achieves the best attack effect in most cases, particularly when the injection ratio is greater than 0.02. Notably, among the recommendation algorithms, APR is more susceptible. This may be caused by the APR algorithm's adversarial property that adds adversarial perturbations during the recommendation model's training phase. Normally, the adversarial term constructed by these perturbations can regularize the model and help the loss function of BPR converge better. Nevertheless, under the attack scenario, the adversarial data (i.e., fake user profiles) naturally cause perturbations, while the adversarial training technique helps the model fit fake users' data. Hence, adversarial training's merit, instead, enlarges the influence of fake users and makes APR more sensitive to shilling attacks. Moreover, GOAT achieves a hit ratio value of more than 10\% on Ciao (multiple) for APR while the injection fraction is 0.05. In this case, for each user, at least one target item hits its recommendation list on average; as a result, GOAT successfully promotes target items to every user.

Additionally, the results also show the potential threats of deep-learning-based shilling attacks (AG and AUSH) and another state-of-art attack (IF) since they often show suboptimal attack effects. It is important to mention that these models sometimes achieve the best attack effect, e.g., AUSH on Douban (multiple) for BPR when the injection fraction is 0.02. However, we also consider that the approximate optimized fake ratings of AG and IF, and AUSH's principle of directly using real user ratings on filler items may impair their attack effects to some degree. We explain for this below.

Although the attacks' effectiveness is achieved, it is clear that attack effects sometimes suffer a loss, i.e., the profit earned from an attack is not always proportional to the injecting fraction. For instance, on Ciao, GOAT suffers such a loss while promoting a single target item in BPR with an injection fraction 0.03; these situations are marked with underlines in Tables \ref{Table3} and \ref{Table4}. The numbers of losses suffered by the models are shown in Table \ref{Table5}, and the results show that GOAT has the lowest probability of suffering a loss.

\begin{table}
\centering
\caption{\textrm{The numbers of losses suffered by the shilling attack models}}
\begin{tabular}{|c|c|c|}
\hline
{} & \multicolumn{2}{c|}{\# of loss suffered} \\ \cline{2-3}
\multirow{-2}{*}{{\begin{tabular}[c]{@{}c@{}}Attack \\ model\end{tabular}}} & {Douban} & { Ciao } \\ \hline
{Ave}  & {0} & {3} \\ \hline
{Ran}  & {1} & {2} \\ \hline
{Band} & {9} & {5} \\ \hline
{UM}   & {2} & {0} \\ \hline
{AG}   & {3} & {3} \\ \hline
{AUSH} & {2} & {3} \\ \hline
{IF}   & {0} & {4} \\ \hline
{GOAT} & {0} & {1} \\ \hline
\end{tabular}
\label{Table5}
\end{table}

For a further analysis, we evaluate the hit ratio ($HR^{'}$@10) of selected items plus filler items among the recommendation lists on BPR. $HR^{'}$@10 is given by Eq. (11), where $ I_S^{'}$ and $ I_F^{'}$ are the sets of selected items and filler items rated by fake users.
\begin{equation}
\begin{aligned}
HR^{'}@10=\frac{1}{|I_S^{'}\cup I_F^{'}|}\sum_{i\in{I_S^{'}\cup I_F^{'}}}\frac{1}{|U|}\sum_{u\in U}\delta{(u,i)}
\end{aligned}
\end{equation}

\begin{figure}[!htbp]
\centering
\subfigure[Douban]
{
\includegraphics{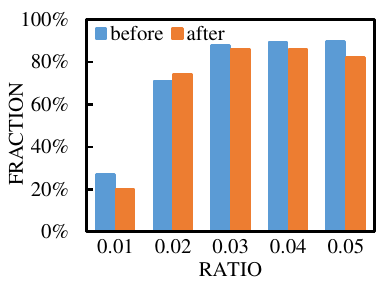}
}
\subfigure[Ciao]
{
\includegraphics{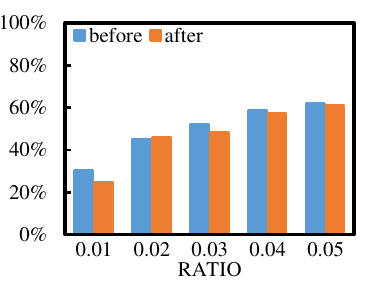}
}
\caption{\textrm{$HR^{'}$@10 on Douban and Ciao}}
\label{Figure11}
\end{figure}

Figure \ref{Figure11} depicts the $HR^{'}$@10 before (blue) and after (orange) GOAT attack. It shows that the selected items and filler items occupy a large portion of the recommendations on both datasets, even before the attacks. On the one hand, these items are naturally recommended by RSs because of their popularity among users; on the other hand, as the injection fraction increases, fake users have more opportunities to cover more candidates for selected items and filler items, promoting these items in RS. Nevertheless, the hit ratio of these items decreases after the attack in most cases because the target items occupy their places, but they still dominate recommendation lists since they outnumber target items. Analytically, if the camouflaged part (i.e., selected items and filler items) in fake user profiles are not sampled and rated properly, their influence can overwhelm the target items' and lead to the decreased attack effects. This explains why Band and AUSH suffer losses on attack effects more frequently -- Band chooses popular items as selected items and gives them the highest ratings, and AUSH samples filler items and their ratings from real user profiles. Both methods magnify the influence of selected items and filler items and are prone to result in such an effect of overwhelming. For AG and IF optimizing the fake ratings by approximated gradients, these gradients are not well regularized, and thus they usually cause extreme ratings on selected items and filler items, i.e., either the highest ratings or the lowest, where the former improves the selected items and filler items and the latter degrades these items. Instead, GOAT leverages a graph convolution structure to smooth fake ratings and avoid extreme ratings to impair attack effects.

\paragraph{\textbf{Attack Cost}}
In this experiment, we compare the attack costs among different models. The attack cost is mainly measured from two aspects: the time cost of fake rating computation and the budget cost of fake rating numbers. For each model, we count the run time and nonzero rating number of generating 100 fake user profiles on the Douban dataset. The models are implemented in TensorFlow and run on a single machine with a GeForce GTX 1080 GPU and the results are shown in Table \ref{Table6}.

These models' strategies of generating attacks can be summarized as i) generating fake ratings for sampled items from an assigned data distribution, e.g., Ave, Ran, Band, UM; ii) generating fake ratings for entire items from a learned data distribution, e.g., AG, IF; iii) generating fake ratings for sampled items from a learned data distribution, e.g., GOAT, AUSH. Table \ref{Table6} indicates that the primitive strategy requires the lowest attack cost since such attacks assume user profiles as Gaussian and directly generates fake ratings from this distribution. Strategies ii) and iii) assume user profiles as more sophisticated distributions, e.g., a network or an optimizer formula, leading to additional computational cost in generating fake ratings. Moreover, strategy ii) generates fake ratings for entire items, i.e., it focuses not only on generating non-zero fake ratings but also zero ratings for unrated items, and in our statistics, each rating value that higher than 0.5 is regarded as a nonzero rating. These attacks cause waste on attack costs since the nonzero rating number usually exceeds the budget of the number of fake ratings and needs to perform downsampling on fake user profiles. Strategy iii) works out a compromise between other strategies; it mainly focuses on generating fake ratings for sampled candidate items, where GOAT's graph convolutional structure consumes extra run time, and AUSH needs to generate extra ratings to align fake user profiles.

In addition, it is important to discuss the scalability of the models. As shown in Table \ref{Table6}, GOAT has the same scalability as primitive models since they only focus on generating fake ratings for sampled candidate items, and the attack cost only relies on the number of fake users and fake ratings that are fully decided by the attacker's intention. However, for other models, the attack costs of generating redundant ratings (zero ratings or fill-in ratings) will grow linearly with increasing size of the item set, making it difficult to extend these methods to a large data scale.

\begin{table}
\caption{\textrm{The attack costs for different attacks}}
\begin{tabular}{|c|c|c|}
\cline{1-3}
Attack model   & Run time (s) & \# of non-zero ratings       \\ \hline
{Ave}          & 0.0041       & 2800                         \\ \hline
{Ran}          & 0.0056       & 2800                         \\ \hline
{Band}         & 0.0072       & 2800                         \\ \hline
{UM}           & 0.0068       & 2800                         \\ \hline
{GOAT}         & 0.1094       & 2800                         \\ \hline
{AUSH}         & 0.0544       & 3400                         \\ \hline
{AG}           & 56.6544      & 47900                        \\ \hline
{IF}           & 0.1089       & 9700                         \\ \hline
\end{tabular}
\label{Table6}
\end{table}

\begin{figure}
\centering
\includegraphics[width=0.25\textwidth]{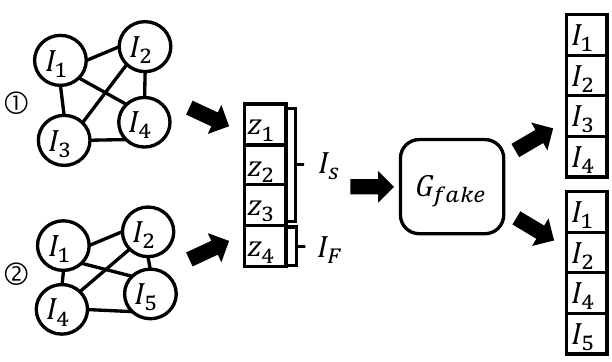}
\caption{An example of generating ratings on the GOAT model}
\label{Figure12}
\end{figure}

\begin{figure}
\centering
\subfigure[Douban]
{
\includegraphics{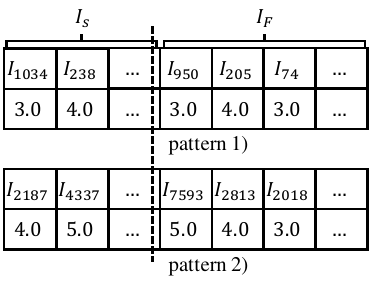}
}
\subfigure[Ciao]
{
\includegraphics{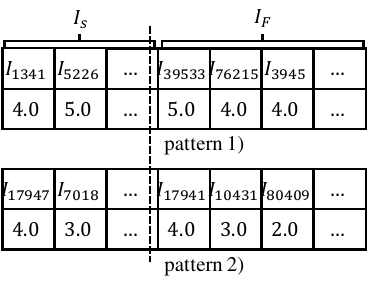}
}
\caption{\textrm{Typical rating patterns on Douban and Ciao}}
\label{Figure13}
\end{figure}

\paragraph{\textbf{What Does Our Shilling Attack Model Learn?}}
It is necessary to mention that each generated rating of GOAT does not correspond to an exact item. Figure \ref{Figure12} shows two cases of generating ratings. In the first case, ratings of $I_3$ and $I_4$ are converted from $z_3$ and $z_4$. In the second case, however, $I_4 $'s rating is converted from $z_3$ rather than $z_4$, and $z_4$ is converted into $I_5$'s rating. The model hereby learns the rating pattern of many sampled items rather than the rating of a single item. The illustration in Figure \ref{Figure13} shows these typical patterns. On Douban, the model learns two patterns of rating distributions. In pattern 1), both selected items and filler items are treated equally, and their ratings float between 3.0 and 4.0. In this case, the fake users do not show their intention regarding the target items specifically, because they want to remain neutral while rating items. By contrast, in pattern 2), selected items tend to have higher ratings than filler items. Thus, they show the preference for target items to some extent. As expected, the graph convolutional structure avoids extreme rating patterns containing either the maximum or minimum rating. The patterns are the same on Ciao as well.

Intuitively, this non-correspondence may be against the goal of simulating the real rating distributions but diversifies the ratings of items by considering different item correlations in different item-item subgraphs. Specifically, in Figure \ref{Figure12}, the correlations between $I_1$, $I_2$, and $I_4$ are different under the two cases, even though they are jointly rated in both cases. Hence, their ratings are different as well.

\paragraph{\textbf{A brief summary}}
The GOAT model generates fake ratings by learning the rating patterns for a set of sampled items. It achieves a expected attack effect while maintaining a low attack cost; i.e., under the same attack size, it pushes the target items into more users' recommendation lists than other attacks. By utilizing the sampling method, the GOAT model can be adjusted adaptively to cope with different data scales and attack costs. From the attacker's view, the attack cost usually is closely coupled with the profits. For the purpose of obtaining profits, attackers would choose deep-learning based attack models along a with multiple-targets scheme, since as long as one target item hits the recommendation lists, their goals are achieved.

\subsection{The Impact of the Attack on Recommendation Performance}
\begin{figure*}
\centering
\subfigure[Douban]
{
\includegraphics{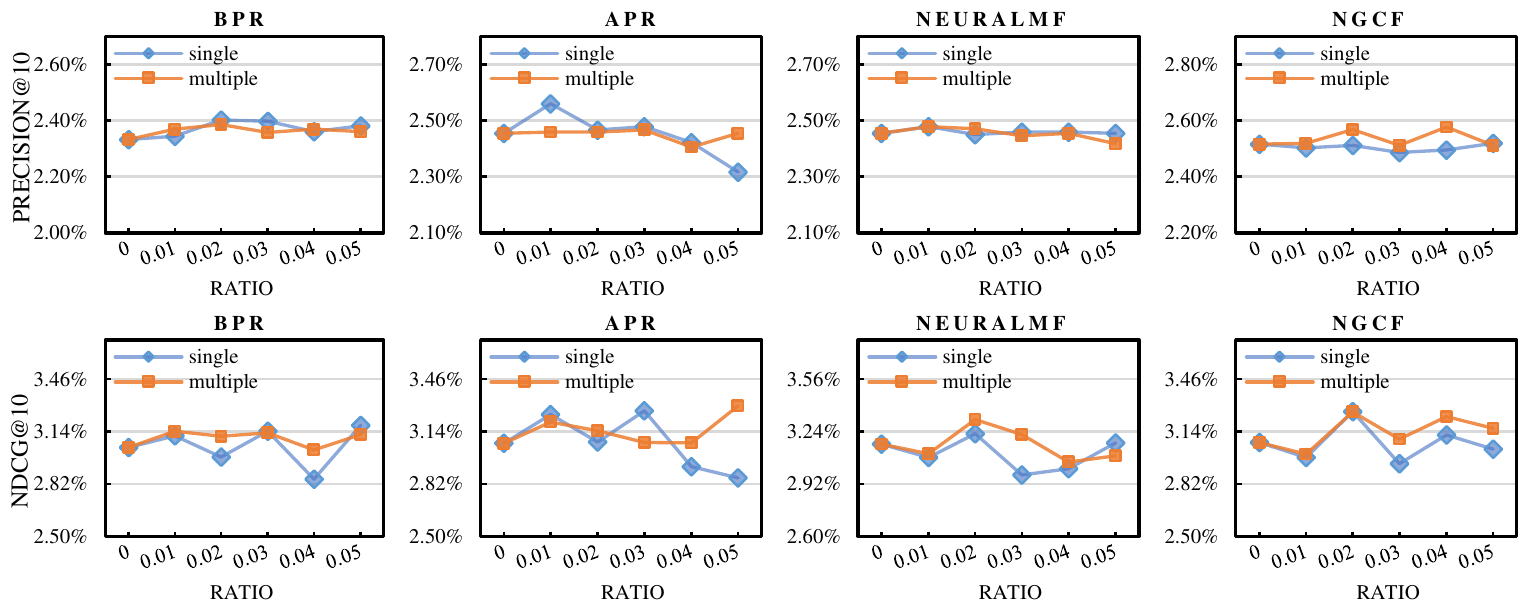}
}

\subfigure[Ciao]
{
\includegraphics{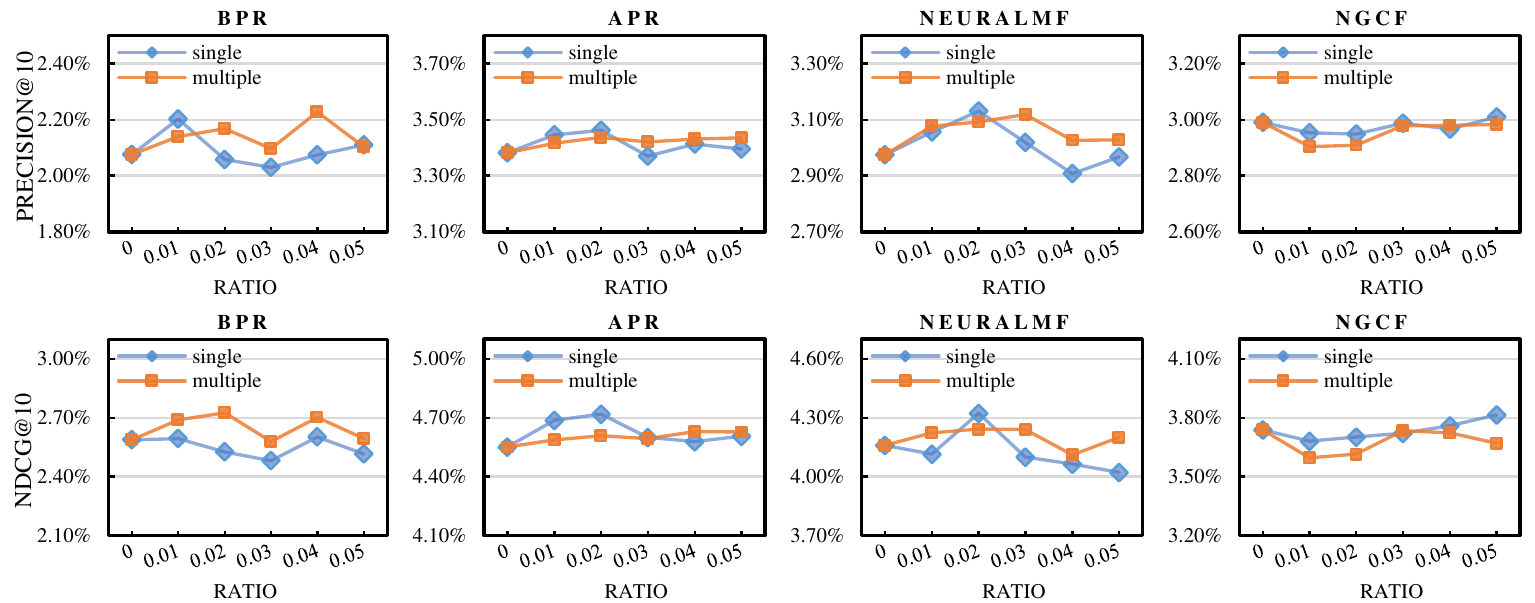}
}
\caption{\textrm{Performance of Recommender Systems Under Different Injection Ratio on Douban and Ciao}}
\label{Figure14}
\end{figure*}

This section specifically addresses the second research question (RQ2) on recommendation performance under attack. The median value of $Precision@10$ and $NDCG@10$ on recommendations are recorded after ten runs of the experiment via the testing set. The results are shown in Figure \ref{Figure14}, where the x-axis is the injection ratio, and the y-axis is the value of a metric. According to the results, the algorithms' performance fluctuates around that without attack. The results show that GOAT does not always deteriorate the recommendation performance even somehow improves it. This may be because the camouflaged part of fake user profiles is learned from real user profiles, and provides more reliable references for recommender systems. In other words, the camouflaged part plays the role of data augmentation\cite{Wang2019a}, which occasionally improves the recommendation performance. This phenomenon also helps fake users disguise themselves since the recommendation performance remains stable in most cases; the founders of RSs can hardly be aware of the system being under attack without specific detection measures.

\paragraph{\textbf{A brief summary}}
The fake user profiles do not mean to deteriorate the recommendation performance but to push up the target item. The camouflaged  part of fake user profiles helps fake users disguise themselves, while the invasive part induces RS to treat the target item equivalent to the recommended one. From the service provider's perspective, such an attack harms the fairness of market competition rather than the performance of RS.

\subsection{Detection of Shilling Attacks}
\begin{figure*}
\centering
\subfigure[Douban]
{
\includegraphics[width=1.5\columnwidth]{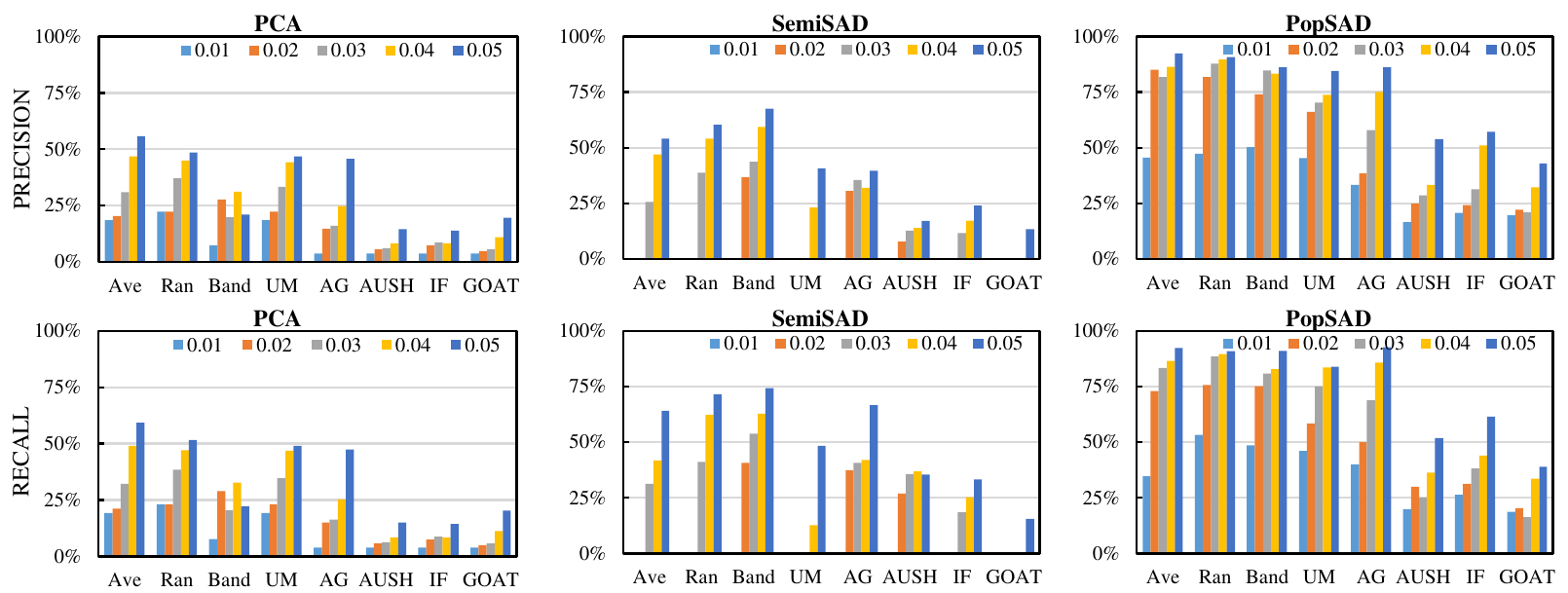}
}
\subfigure[Ciao]
{
\includegraphics[width=1.5\columnwidth]{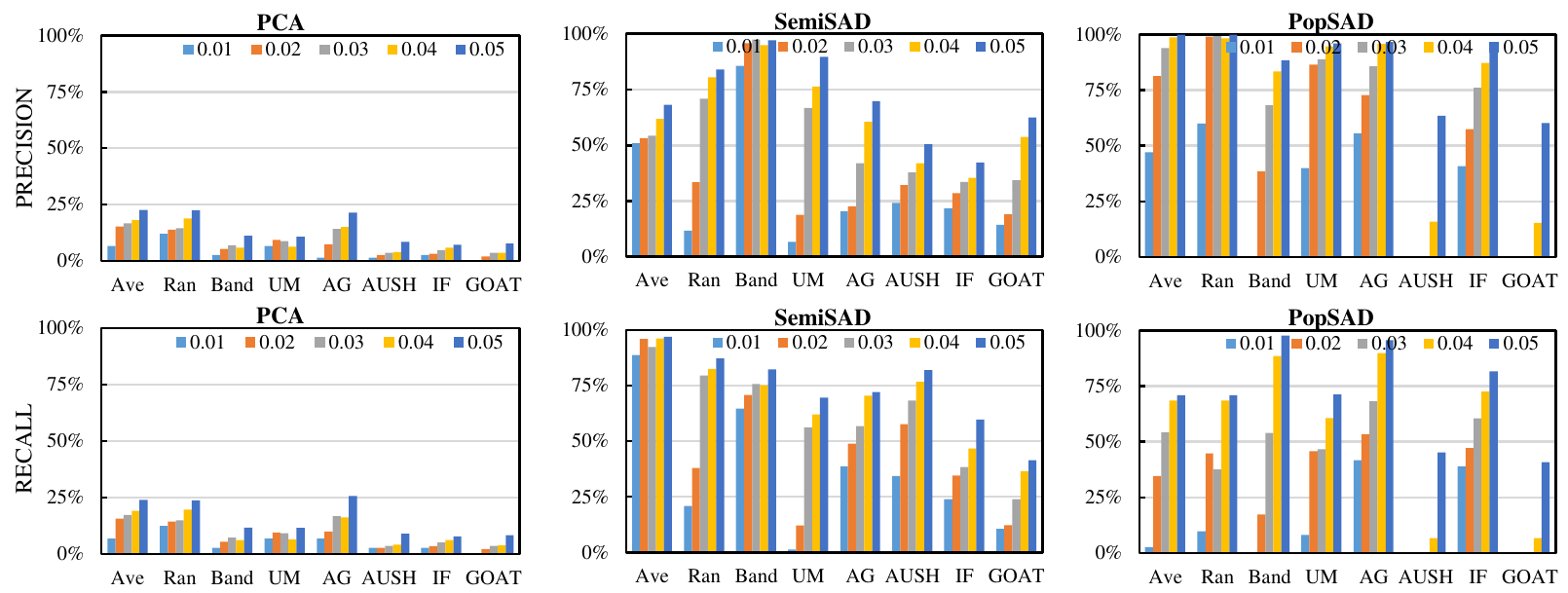}
}
\caption{\textrm{Detection Performance Under Different Injection Ratios on Douban and Ciao}}
\label{Figure15}
\end{figure*}

This section addresses the third research question (RQ3) on shilling attack detection. Three shilling attack detection algorithms -- PCASelectUsers \cite{Mehta2009}, SemiSAD \cite{Cao2013} and DegreeSAD \cite{Li2016a} are used to detect fake users (under the 'multiple' attack setting). These methods are selected as representative unsupervised, semi-supervised, and supervised shilling attack detection algorithms. Specifically, the unsupervised method is run ten times and takes the median value as a result; Semi-supervised and supervised methods are run with fivefold cross-validation to obtain the results, and there are 50\% labeled data for semi-supervised method training. The results are shown in Figure \ref{Figure15}, where the x-axis indicates the injection ratio of different attacks, and the y-axis is the value of the metric. The results only focus on the detection performance of fake users because the injection ratios are relatively small. For example, consider the case of one hundred users in total but only five fake users. Even if the detector regards all users as authentic, it still reaches a 95\% accuracy, which is unreasonable since it fails to detect the fake users. Additionally, it needs to be mentioned that both metrics on classifying real users are above 95\%.

From the attacker's view, deep-learning-based (AUSH and GOAT) attacks are the most undetectable under the three detection methods. These models learn more accurate rating distributions to disguise fake data, and it is more difficult for the detectors to distinguish fake users from real users than for the traditional handcrafted attacks. As mentioned above, AG and IF attacks usually generate extreme ratings that are not conducive to the camouflage of fake users; hence, they are easier to detect than AUSH and GOAT. Notably, under the limited attack cost setting, even the traditional shilling attack models can escape detection, e.g., when the injection ratio is 0.01, the SemiSAD method on Douban fails to detect fake users, and the PCA method can hardly detect fake users on Ciao under all attack schemes and injection ratios. However, the attacker's malice will be exposed gradually as the injection ratio increases. Although AUSH shows similar camouflage ability to GOAT, it is still exposed earlier than GOAT as the injection ratio increases in some cases (see SemiSAD on Douban while the injection ratio ranges from 0.02 to 0.04, and PCA on Ciao while the injecting ratio is 0.01). As a result, the increasing injection ratio means not only more profits but also more risks of being detected.

From the detector's view, one of the challenges they face is obtaining the labeled user data. The unsupervised method is a good solution to this problem, but it also has some drawbacks. Based on the results presented in Figure \ref{Figure15}, we argue that the unsupervised method has low effectiveness to detect fake users compared with other detection methods. Although the semi-supervised and supervised methods are superior to the unsupervised method, the small injection ratio results in an imbalance between fake labels and real labels that may cause underfitting or overfitting problems. For instance, the SemiSAD method on Douban loses its detection ability when the injection ratio is 0.01 and the PopSAD method on Ciao loses its detection ability on Band when the injection ratio is 0.01 as well as on AUSH and GOAT when injecting ratio ranges from 0.01 to 0.03.

\paragraph{\textbf{A brief summary}}
The detection results show that even GOAT cannot always ensure the disguise ability, and the fake users will be exposed at last as the injection ratio increases. From the detector's perspective, to achieve a favorable trade-off between the cost of labeling data and the performance of detections, the semi-supervised method is preferred.

\subsection{Defensive Strategies of GOAT}
This section answers to the fourth research question (RQ4) regarding the possible prevention and detection measures for counterattack. Through the above experimental analysis, we suggest the following prevention and detection measures of GOAT.

\textbf{Anonymous Data.} Anonymization of the data appears to be most effective against potential attacks. The GOAT model utilizes average item ratings and user historical behaviors to generate fake user profiles. The rating information is usually disclosed in the system to describe item qualities, but the users can be anonymous so that the attacker cannot track user historical behaviors to elaborate fake user profiles.

\textbf{Pre-detection.} Pre-detection is another strategic measure. The unsupervised detection can be deployed as a pre-detection strategy to verify whether the recommender system is under attack since it is not affected by the imbalanced label. The service provider can employ specialists to identify the detected fake users to be aware of the attack and then deploy more advanced detectors.

\section{Conclusion and Future Work}
In this work, we devise a novel model of GOAT that combines GAN and GNN to deploy a compromised shilling attack that solves the contradiction between the cost and effectiveness. Our attack can be generalized as sampling and rating items for fake users. The sampling method is adopted to confine the attack cost. To overcome the model collapse problem of GAN training, we separate the whole training in stages, propose a rating penalty term to the generator loss function, and add extra information to the input GAN. We also use a tailored graph convolution structure that models the high-order co-rated item relationship to smooth the fake ratings. Through evaluations on two public datasets, we find that a deep learning-based attack can inflict more harm on the CF-based recommender systems even at limited costs. We also analyze the attack effect in terms of attackers, service providers, and detectors to provide advice for the prevention and detection of such an attack. The prevention measure aims to prevent the attackers from obtaining access to authentic data, while the detection measure helps the service provider perceive the occurrence of an attack and identify malicious users. 

This work mainly explores the vulnerability of CF-based RSs; however, other recommendation approaches, such as content-based RSs and hybrid RSs are not mentioned. Hence, a possible future direction of this research is to extend this methodology to various types of RSs. Further studies on shilling attacks could alert researchers and service providers to the vulnerability of RSs. Multiple defense mechanisms must be considered in the deployment of a recommender system.

\section*{Acknowledgements}
This study was supported by the National Key Research and Development Program of China (2018YFB1403602),  the National Natural Science Foundation of China (71532002), the Natural Science Foundation of Chongqing, China (cstc2020jcyj-msxmX0690), the Overseas Returnees Innovation and Entrepreneurship Support Program of Chongqing (cx2020097), the Fundamental Research Funds for the Central Universities of Chongqing University (2020CDJ-LHZZ-039),  and also partially supported by the Technological Innovation and Application Program of Chongqing (cstc2019jscx-zdztzxX0031).

\bibliographystyle{cas-model2-names}


\printcredits
\vskip 3pt

\end{document}